\pdfoutput=1

\documentclass[11pt]{article}
\usepackage{EMNLP2022}

\usepackage{times}
\usepackage{latexsym}

\usepackage{subfig}
\usepackage{multirow}
\usepackage{algorithm}
\usepackage{algorithmic}
\usepackage{graphicx}

\usepackage[T1]{fontenc}

\usepackage[utf8]{inputenc}

\usepackage{microtype}

\title{Capturing Topic Framing via Masked Language Modeling}
\author{Xiaobo Guo, Weicheng Ma \and Soroush Vosoughi \\
         Department of Computer Science, Dartmouth College\\
         Hanover, New Hampshire\\
         \{xiaobo.guo.gr, weicheng.ma.gr \and soroush.vosoughi\}@dartmouth.edu
}

\begin{document}
\maketitle
\begin{abstract}
    Differential framing of issues can lead to divergent world views on important issues. This is especially true in domains where the information presented can reach a large audience, such as traditional and social media. Scalable and reliable measurement of such differential framing is an important first step in addressing them. In this work, based on the intuition that framing affects the tone and word choices in written language, we propose a framework for modeling the differential framing of issues through masked token prediction via large-scale fine-tuned language models (LMs). Specifically, we explore three key factors for our framework: 1) prompt generation methods for the masked token prediction; 2) methods for normalizing the output of fine-tuned LMs; 3) robustness to the choice of pre-trained LMs used for fine-tuning. Through experiments on a dataset of articles from traditional media outlets covering five diverse and politically polarized topics, we show that our framework can capture differential framing of these topics with high reliability.

\end{abstract}
\section{Introduction} \label{sct:introduction}

    Issue framing refers to the ways in which information organizations (e.g. media companies, government institutions, etc.) package and present information \cite{nelson2011issue}. The framing of the information is of great importance because it influences readers' understanding of the content \cite{kamoen2019issue, carreras2021does}. Issue framing can be expressed explicitly with evaluative language (e.g., ``good'' or ``bad'') or implicitly (e.g., ``undocumented'' vs ``illegal" immigrants) \cite{jacoby2000issue,zhang2021media}. Measuring the ways in which topics are framed differently (what we term ``differential framing'') is helpful in identifying the various stances of these information organizations.
    
    While much research has been done on qualitative analysis and the effects of issue framing (e.g., see \cite{jacoby2000issue, sniderman2004structure,jerit2008issue,zhang2021media}). Quantitative methods have been used for this task, though these methods usually utilize hand-crafted features such as the tone or the choice of phrases when referring to the same events or entities (e.g., undocumented immigrant vs. illegal alien). These methods either rely fully on human annotation \cite{matthes2009s,chinn2020politicization,lim2018understanding,golez2020home,farber2020multidimensional,gentzkow2010drives,fan2019plain}--which makes them not scalable and prone to human bias-- or utilize (semi-) automatic methods by representing the problem as a standard supervised learning task \cite{spinde2020integrated,spinde2020media,spinde2021automated,chen2020analyzing}, relying on context-specific hand-crafted features.
    
    To address the challenges of scale and topic-/context-specificity, inspired by prior works on prompt-based knowledge extraction\cite{petroni2019language,roberts2020much,qin2021learning,perez2021true,guo2022measuring} which shows that knowledge learned by LMs can be extracted through the masked language modeling (MLM) task, we propose a novel method that utilizes LMs to measure differential framing in any context or domain. Specifically, we capture the tone and word preference of different sources by fine-tuning pre-trained LMs on corpora from them. Through the MLM objective, we then use these fine-tuned LMs to predict the choice of words for each source for different contexts, such as different issues or events. In other words, prompting these models with sentences such as ``the greatest threat to the immigration system is $\_\_\_$'', they will output the word most likely to complete the sentence given the associations learned from their fine-tuning texts. Comparing outputs from models fine-tuned on different sources can illuminate key differences in their framing of issues.
    
    We explore three key factors in the development of our framework: 1) how to generate the prompts; 2) how to normalize the output of fine-tuned LMs; 3) robustness of our method to different LMs. In this paper, we explore differential framing in media outlets as a test case. Specifically, we capture the tone and word preference of media outlets by fine-tuning LMs on corpora from different outlets and topics collected from Media Cloud, a publicly available news API. The difference in framing between outlets is then used to rank the outlets based on their similarity in framing. Our rankings are validated through human evaluations and comparisons with three ground truth datasets.
    
    Our proposed framework for estimating differential framing is highly scalable and generalizable in that it can be applied to other contexts and domains with minor adjustments. We make our code and data publicly available on github\footnote{https://github.com/guoxiaobo96/media-position}.

\section{Dataset Construction}

    Our dataset is comprised of articles in English collected from Media Cloud \footnote{https://mediacloud.org/}, whose topic-labeling has proved reliable \cite{etling2014blogs}. Specifically, we sample news articles under five diverse topics: ``Climate Change'', ``Corporate Tax'', ``Drug Policy'', ``Gay Marriage'' and ``The Affordable Care Act''. The articles are from 10 US news outlets: Breitbart News Network, CBS News, CNN, Fox News, HuffPost, New York Times, NPR, USA Today, Wall Street Journal, and Washington Post. 
 
    We divide the dataset into training and development (dev) sets with a 90/10 split.
    Due to the length limit of LMs, we break the news articles down to their original paragraphs; paragraphs exceeding 256 words are split greedily at the sentence level to ensure the completeness of sentences. We then label the paragraphs with their respective media outlets. In the final dataset, there are 531,398 instances. The statistics of these datasets are listed in Appendix \ref{sct:dataset} (Table \ref{tab:data}).

 \section{Methodology}\label{sct:methodology}
    
\begin{figure*}[!ht]
        \centering
        \includegraphics[width=2.05\columnwidth]{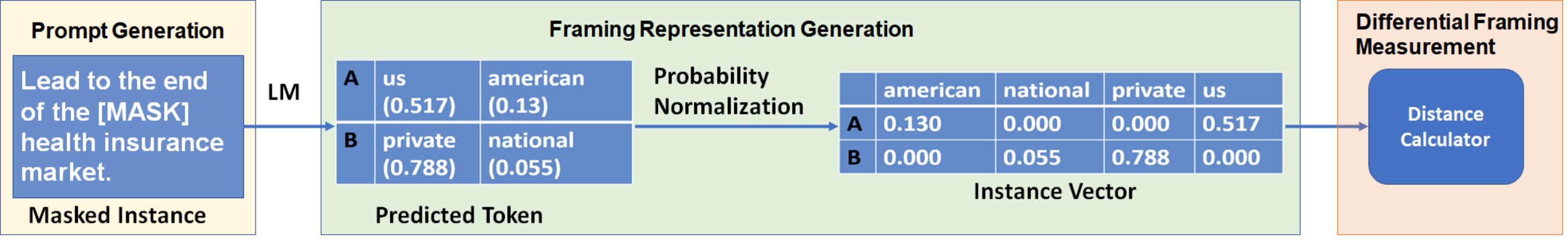}
        \caption{An example generation of representations for two sources. For simplicity, we generate the representations using the top-2 predicted words and the no-normalization method.}
        \label{fig:encoder}
        \vspace{-15pt}
    \end{figure*}

    As shown in Figure \ref{fig:encoder}, with the fine-tuned LMs for each source, our framework is comprised of three steps: (1) we generate prompts for the MLM task; (2) with the given prompts, we then leverage the fine-tuned LMs to create representations of the sources via prompt-based mask token prediction; and (3) we utilize the generated representations to measure differential framing by the sources. We experiment with three different fine-tuned models (using the MLM task), ``RoBERTa-base'' \cite{liu2019roberta}, ``BERT-base-cased'' and ``BERT-base-uncased'' \cite{devlin2019bert}. 

    \subsection{Prompt Generation}
        For the task of prompt-based mask token prediction, the prompt generation method will strongly influence its performance \cite{petroni2019language,chen2021adaprompt,shin2020autoprompt,ben2021pada}. In this section, we explore automatic and manual methods used for prompt generation.
        
        We explore different automatic methods for selecting prompts and masked tokens (see Appendix \ref{sct:mask-model}). The most effective method (called ``bigram outer'' works as follows: inspired by work on authorship attribution (e.g., \cite{coyotl2006authorship}), we rely on the words all sources have in common. Specifically, we first create a list of bigrams that appear in the dev set of all sources. For each instance in the dev set containing these bigrams, we generate two masked prompts (per bigram), one where the mask is applied to the word preceding the bigram, and one with the mask applied to the word following the bigram. 
        
\begin{figure}[!ht]
        \centering
        \includegraphics[width=1.00\columnwidth]{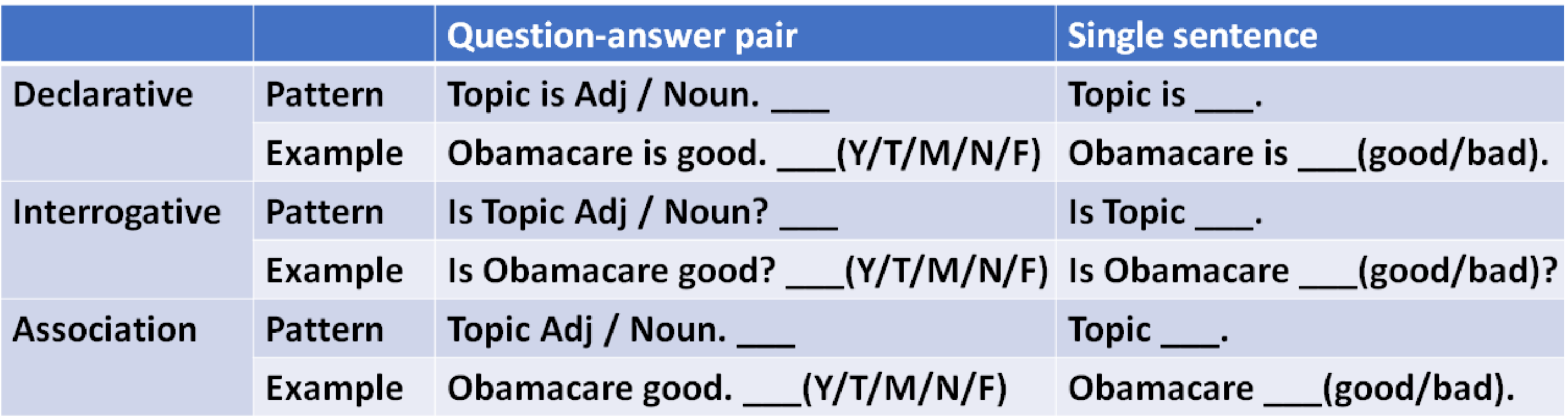}
        \caption{Examples of manually generated prompts. Here, the topic word is ``Obamacare'', and the Adjective is ``good''. ``\_\_\_'' is the masked token, and the tokens in the bracket are the choices for prediction.``Y'':``Yes'' , ``T'':``True'',``M'':``Maybe'', ``N'':``No'', ``F'':``False''.}
        \label{fig:pattern}
        \vspace{-15pt}
    \end{figure}

        We also explore different manually generated prompt patterns based on domain knowledge. Each pattern includes topic words (``Topic''), an adjective (``Adj'') or a noun with its corresponding indefinite article (``Noun''), and a ``To Be'' auxiliary verb. We evaluate declarative (``Topic is Adj/Noun.''), interrogative (``Is topic Adj/Noun?''), and association patterns (``Topic Adj/Noun.'' or ``Adj/Noun Topic.''). We also explore question-answer and single sentence patterns. For the former, we keep the whole sentence and add a masked token after the sentence to predict affirmative, unsure, and negative responses. For the latter, we mask the ``Adj'' or the ``Noun'' to be predicted. Different from the automatic method, here we limit the possible choices for the predicted tokens. For the question-answer pattern, the choice of predicted token is limited to ``Yes'', ``True'',``Maybe'', ``No'', and ``False''. For the single sentence pattern, we limit the choice to a pair of antonyms showing the attitude (e.g. ``good'' vs ``bad''). Figure \ref{fig:pattern} shows examples of these patterns.
        
    \subsection{Framing Representation Generation}
        With the generated prompts, we then use the fine-tuned LMs for each source to predict the masked tokens. Considering that the probability of each token is different in the non-fined-tuned pre-trained LMs, we explore two different methods for normalizing the predicted probabilities. The first (called general-normalization) divides the predicted probability of a token from the fine-tuned LM by its predicted probability from the non-fine-tuned LM. Similarly, the second method (called domain-normalization) normalizes using a domain-specific LM which is fine-tuned on all instances in the training sets. 
        
        For the automatically generated prompts, we retrieve the top-K candidate words with the highest probability and, for manually generated prompts, we retrieve the probabilities of the pre-selected tokens. The differential framing of each source with respect to the masked prompt is then represented as a vector of the probability of these words. Note that the vectors for the sources all have the same length and correspond to the same words, i.e., the union of all top-K candidate words from the sources. If a word was not in the top-K candidate of a source, its probability is set to 0. This allows for cross source comparison. Figure \ref{fig:encoder} shows an example of how the differential framing representations are generated for two sources using the top-2 candidate words.
        
        For each topic $t$ we have $n_t$ vectors for each source, where $n_t$ is the number of masked prompts for each topic. The set of these vectors represents the differential framing of each source with respect to different topics. 

    \subsection{Differential Framing Measurement}
    \label{mrb}
        We measure the differential framing by the sources within specific topics by first calculating the distance between each pair of sources. To do so, we calculate the mean of the cosine distance across all aligned differential framing representation vectors for the specified topics across each source pair. 
        
        Next, based on their distance, we create a similarity ranking for each source. Note that these rankings do not have to be symmetric, in that if source A has source B as the closest source, it does not necessarily follow that B will have A as its closest source. These rankings indicate the differential framing similarity of each source with respect to the others.
        Similar rankings suggest similar framing of the specified topics by the sources and vice versa.
        
\section{Experiments \footnote{Details for reproducibility are shown in Appendix \ref{sct:expe}.}}
    We explore and evaluate our framework in the context of differential framing of topics by different media outlets. 
    Appendix \ref{sct:dataset} describes the construction and details of the media outlets data used. 
    We evaluate the performance of our framework through human evaluations and three survey-based datasets. Appendix \ref{sct:human-eval-data} describes the collection of our human evaluation dataset. The construction of the other three survey-based datasets is described in Appendix \ref{sct:data-g}. We compare the similarity rankings of the outlets predicted by our framework with the ones generated using the four datasets described above. Specifically, we measure the relative rankings of the outlets based on our ground truth datasets using a similar procedure as described in Section \ref{mrb}: for each outlet, we calculate the absolute distance to the other outlets (based the ground truth data) and create a ranked list. We then calculate the similarity between the predicted and ground truth similarity rankings via Kendall rank correlation coefficients (Kendall's $\tau$). Kendall's $\tau$ is calculated for each media outlet, and the mean of all the $\tau$'s is used as the evaluation score. 

    Here, we report the human evaluation results. Table \ref{tab:pattern-stats} shows the influence of prompt generation methods. As can be seen, for all manual methods, the single sentence achieves better performance than the question-answer pair pattern (average $\tau$ of 0.43 vs 0.34), showing that adjective or noun prediction is a better choice compared to polar answers.
    The fact that ``decalarative'' (avg ($\tau$) = 0.40) and ``interrogative'' (avg ($\tau$) = 0.40) perform better than ``association'' (avg ($\tau$) = 0.37) shows that sentence structure, regardless of its type, is helpful for token prediction.
    The best performing manual pattern, ``declarative-single'' (0.46), achieves slightly better performance than the automatic ``bigram outer'' (0.44), showing that manually designed prompts are slightly better than automatic ones, presumably due to their careful design. 
    Overall, the standard deviation of the manual methods is higher across the board, which might be due to the limited number of manually designed prompts.
    \begin{table}[hbt]
    \centering
        \begin{tabular}{llll}
        \hline
                                       &        & mean & std  \\ \hline
        \multirow{2}{*}{declarative (manual)}   & q\&a   & 0.34 & 0.07 \\
                                       & single & \textbf{0.46} & 0.07 \\ \hline
        \multirow{2}{*}{interrogative (manual)} & q\&a   & 0.35 & 0.14 \\
                                       & single & 0.44 & 0.07 \\ \hline
        \multirow{2}{*}{association (manual)}   & q\&a   & 0.34 & 0.06 \\
                                       & single & 0.39 & 0.11 \\ \hline
        bigram outer (auto)                   &        & 0.44 & 0.06 \\ \hline
        \end{tabular}
        \caption{Agreement ($\tau$) between media similarity rankings using our framework and the human evaluation dataset. ``q\&a'' means question-answer pair pattern, and ``single'' means single sentence pattern.}
        \label{tab:pattern-stats}
        \vspace{-10pt}
    \end{table}
    
    Table \ref{tab:norm-stats} shows the mean performance of each normalization method. Both general-normalization (0.42) and domain-normalization (0.44) achieve much better performance than no-normalization (0.24). This shows that it is important to normalize the probabilities generated by the LMs as it maximizes the differences between sources.

    \begin{table}[!hbt]
        \centering
        \begin{tabular}{l|ll}
        \hline
                              & mean & std  \\\hline
        no-normalization     & 0.22 & 0.11 \\
        general-normalization & 0.42 & 0.09 \\
        domain-normalization  & 0.44 & 0.07 \\\hline
        \end{tabular}
        \caption{Agreement ($\tau$) between media similarity rankings using our framework and the human evaluation dataset.}
        \label{tab:norm-stats}
        \vspace{-10pt}
    \end{table}
    
    Table \ref{tab:lm-stats} shows the performance of ``Roberta-base'' (0.40), ``BERT-base-uncased'' (0.38),  and ``BERT-base-cased'' (0.40). The choice of LM seems to have little effect on our framework.
    \begin{table}[ht]
    \centering
    \begin{tabular}{l|ll}
    \hline
                      & mean & std  \\\hline
    RoBERTa-base      & 0.40 & 0.09 \\
    BERT-base-uncased & 0.38 & 0.09 \\
    BERT-base-cased   & 0.40 & 0.10 \\\hline
    \end{tabular}
        \caption{Agreement ($\tau$) between media similarity rankings of our framework and the human evaluation dataset.}
        \label{tab:lm-stats}
        \vspace{-10pt}
    \end{table}
    
    Appendix \ref{sct:human-eval-stats} shows the performance of all different combinations for the human evaluation dataset. Appendix \ref{sct::auto-results} shows the results of the same experiments on the other three datasets.

    Finally, we compare the performance of our framework against four baselines. For our framework, we use the combination of three factors achieving the best performance: ``interrogative'', ``one sentence'', and domain-normalization. 
    Two classic baselines are based on Latent Dirichlet allocation (LDA) and Term Frequency-Inverse Document Frequency (TF-IDF) features. The other two baselines utilize language models (``RoBERTa-base'', ``BERT-base-cased'', or ``BERT-base-uncased''), with one using the classification objective (LM-c), a variant of  previous work\cite{chen2020analyzing}, and the other using the MLM objective (LM-m). The LDA and TF-IDF (up to trigrams) are trained with all media outlet instances and utilized to embed the instances. Following previous work, LM-c is fine-tuned with the task of predicting the source of each instance and leveraged to embed all instances in the dev sets. For LM-m, we use our MLM fine-tuned language models to embed the instances in the dev sets. In all cases, we regard the mean embedding of all instances for one media outlet as its corresponding media embedding.  

    For all the baselines with the media outlet embeddings, we calculate the cosine distances between pairs of media outlets in a manner similar to what was described in Section \ref{mrb}, i.e., for each outlet, we use the distance to other outlets to create a ranked list and then use Kendall's $\tau$ to measure agreement with the ground truth rankings. Appendix \ref{sct:hyper-and-seed} describes the details of the hyper-parameters of the baselines. Table \ref{tab:media_bais_com} shows the mean $\tau$ across all topics. Our framework outperforms all baselines. Comparing the performance of our methods and LM-m, we see that predicting masked tokens can achieve better performance than sentence embedding.

\begin{table}[!hbt]
\vspace{-5pt}
\centering
\begin{tabular}{l|lll}
 \hline
       & RoBERTa             & BERT-un             & BERT-ca             \\ \hline
LDA    & 0.34(0.09)          & 0.34(0.09)          & 0.34(0.09)          \\
TF-IDF & 0.33(0.04)          & 0.33(0.04)          & 0.33(0.04)          \\
LM-c   & 0.30(0.11)          & 0.44(0.10)          & 0.49(0.08)          \\
LM-m   & 0.37(0.12)          & 0.46(0.14)          & 0.43(0.13)          \\
ours   & \textbf{0.39(0.15)} & \textbf{0.50(0.18)} & \textbf{0.53(0.08)} \\ \hline
\end{tabular}
    \caption{Agreement ($\tau$) between the human evaluation dataset and the different methods. Results are averaged across 5 topics. Stds are shown in parentheses. }
    \label{tab:media_bais_com}
    \vspace{-15pt}
\end{table}

\section{Error Analysis}
        
    To better understand what our model is capturing, we rank the differential framing for each masked instance and calculate the Kendall's $\tau$ based on the instance-level rankings.  We observe that all distributions are right-skewed distributions with a long right tail (see figure \ref{sct::error-analysis} in Appendix \ref{sct::error-analysis}).

    We further analyze the instances with the worst and best performance of each topic and ground truth dataset. We observe that the instances with the worst performances (i.e., small $\tau$), have masked tokens that are easy to predict by all models no matter which articles they were fine-tuned on, by either being a stop word or a named-entity (such as ``Nancy Pelosi''). For example, in the following sentence: ``... age-based tax credits in the bill replacing those in $\_\_\_$ affordable care act are too small'', the masked token is ``the'', and all models correctly predict it with the probability above 0.99.  
        
   We also analyze the instances which achieve the best performance. As hypothesized, in many of the instances the difference in the prediction can be explained by the difference in opinion of the outlets. For example, in the sentence "President Donald Trump's vague platitudes about beautiful and very $\_\_\_$ health care for everybody...'', the masked token is special. For the right-wing media outlets (such as Fox and Breitbart), the models tend to predict positive tokens (such as ``effective'' and ``affordable''), and for the left-wing media outlets (such as CNN and New York Times), the models will predict both positive (``affordable'') and negative (``expensive'') tokens.

\section{Conclusion and Future Work}
    We present a framework for capturing differential framing of issues using the MLM objective through masked prompting of fine-tuned large-scale LMs. We explore three factors: prompt generation methods, probability normalization, and robustness to different LMs. We show that manually designed prompts can achieve slightly better performance than the automatically generated prompts at the cost of volatility. We also show that the normalization of outputs from the fine-tuned LMs can greatly boost the performance of our framework. Finally, we show that our framework is robust to the choice of LMs. We evaluate our framework on differential framing of issues by media outlets, using human evaluations and three survey-based datasets, showing that our framework's predicted differences in the ways in which outlets frame issues mainly agree with these ground truth datasets. Future work could investigate the applicability of our framework to other domains.

\section{Limitations}
    There are four main limitations to our work. The first is the lack of interpretability for both the manual and automatic prompts. For the manual prompt, the  predicted tokens are sometimes unrelated to the prompt, which is why we limit the set of possible tokens to be predicted. More work is needed to understand the cause behind the prediction of such tokens. Similarly, the prompts and the masked tokens generated by the automatic method are also sometimes not relevant to the issue or topic being investigated.   
    The second limitation of our work is that all experiments have been conducted in English (mainly due to a lack of appropriate ground truth data in other languages). We do not currently know whether our framework generalizes to other languages. Further experiments are needed for this. 
    Third, again due to a lack of ground truth data, our paper only investigates our framework in the context of differential framing of issues by different media outlets. Though our framework is designed to be domain-agnostic, we currently do not have the experimental results to show this.
    Finally, our framework measures word-based differential framing. There are other types of differential framing (i.e., discourse-level) that our framework cannot currently capture.
    
\section*{Acknowledgements}
We sincerely thank the reviewers for their insightful comments and suggestions that helped improve the paper. This research was supported in part by a Google Research Scholar Award.

\bibliography{main}
\bibliographystyle{acl_natbib}
\clearpage
\setcounter{table}{0}
\setcounter{figure}{0}
\renewcommand\thefigure{\Alph{section}\arabic{figure}}
\renewcommand\thetable{\Alph{section}\arabic{table}}
\appendix

\section{Dataset Details}
\label{sct:dataset}
        The statistics of the train and dev sets for each topic and media outlet are listed in Table \ref{tab:data}.

    \begin{table*}[h!bt]\small
        \centering
\begin{tabular}{l|rr|rr|rr|rr|rr}
            \hline
          & \multicolumn{2}{r|}{Climate Change}                     & \multicolumn{2}{r|}{Corporate Tax}                      & \multicolumn{2}{r|}{Drug Policy}                        & \multicolumn{2}{r|}{Gay Marriage}                       & \multicolumn{2}{r}{The Affordable Care Act}                          \\
          & \multicolumn{1}{r}{Train} & \multicolumn{1}{r|}{Dev} & \multicolumn{1}{r}{Train} & \multicolumn{1}{r|}{Dev} & \multicolumn{1}{r}{Train} & \multicolumn{1}{r|}{Dev} & \multicolumn{1}{r}{Train} & \multicolumn{1}{r|}{Dev} & \multicolumn{1}{r}{Train} & \multicolumn{1}{r}{Dev} \\
          \hline
Breitbart  & 8,267                        & 807                      & 8,357                        & 756                      & 7,765                        & 934                      & 7,663                        & 1,016                    & 6,665                        & 732                     \\
CBS        & 11,515                       & 1,680                    & 9,002                        & 1,026                    & 9,313                        & 1,273                    & 12,320                       & 2,268                    & 7,968                        & 799                     \\
CNN        & 12,587                       & 1,414                    & 12,982                       & 1,529                    & 13,158                       & 1,565                    & 13,949                       & 1,841                    & 16,030                       & 1,805                   \\
Fox        & 8,487                        & 450                      & 8,526                        & 1,284                    & 9,445                        & 602                      & 6,473                        & 695                      & 7,272                        & 735                     \\
HuffPost   & 10,272                       & 1,102                    & 11,044                       & 1,087                    & 9,780                        & 1,117                    & 9,385                        & 1,054                    & 10,138                       & 1,132                   \\
NPR        & 15,260                       & 1,509                    & 14,730                       & 1,997                    & 14,934                       & 1,503                    & 17,285                       & 2,036                    & 13,057                       & 1,281                   \\
NYTimes    & 3,113                        & 322                      & 2,678                        & 297                      & 3,077                        & 271                      & 3,803                        & 289                      & 4,936                        & 452                     \\
USA Today   & 12,288                       & 1,842                    & 13,464                       & 1,476                    & 12,432                       & 1,718                    & 12,436                       & 1,340                    & 12,261                       & 1,208                   \\
Wallstreet & 4,914                        & 447                      & 4,040                        & 433                      & 4,403                        & 469                      & 6,469                        & 635                      & 3,003                        & 407                     \\
Washington & 9,747                        & 1,098                    & 8,788                        & 816                      & 9,545                        & 911                      & 7,666                        & 861                      & 14,459                       & 1,947   \\
\hline
\end{tabular}
        \caption{The statistics of the train and dev sets for each topic and media outlet.``NYTimes'': ``New York Times'', ``Wallstreet'': ``The Wall Street Journal'', ``Washington'': ``Washington Post''}
        \label{tab:data}
    \end{table*}

\section{Prompt Generation}
\setcounter{table}{0}
    \subsection{Mask Method}
        \label{sct:mask-model}
        For the prompt-based mask token prediction, the choice of prompts and the word to be masked has a significant effect on the quality of generated media attitudinal representations. We explore five automatic methods for selecting prompts and words to mask:
        
        \textbf{Random sampling (RS)} is the most basic method. We choose 50\% of the instances uniformly at random and replace 10\% of the words in each of them with the special token [MASK]. In our experiments, we regard the random sampling method as the baseline for evaluating the influence of masked tokens and prompt selection methods.
        
        \textbf{BERT attention (BERT)} chooses the words to mask based on the attention scores generated by BERT. Specifically, for each topic, we train a classifier based on bert-base-cased to predict the label (i.e., the media outlet) of each instance using the train set. We then apply the classifier to predict the label of instances in the dev set of the same topic and extract the attention scores from the twelfth layer of the trained classifier as the importance measure of each token. We perform word-masking only on the instances whose labels are correctly predicted by the model with confidence scores above 0.7 \footnote{We choose 0.7 because our models achieve the best performance in our small-scale experiments with this threshold}. For each selected instance, we mask the token with the highest importance score. 
        
        \textbf{Bigram inner (BI)} works as follows: (1) We first create a list of bigrams that appear in the dev set of at least five media outlets (2) For each instance in the dev set containing these bigrams we generate two masked prompts (per bigram), one where the mask is applied to the first word of the bigram and one with the mask applied to the second word of the bigram. 

        \textbf{Trigram inner (TI)} is similar to \textbf{bigram inner} while considering trigrams instead of bigrams.
        
        \textbf{Bigram outer (BO)} works as follows: (1) We first create a list of bigrams that appear in the dev set of all ten media outlets (2) For each instance in the dev set containing these bigrams we generate two masked prompts (per bigram), one where the mask is applied to the word preceding the bigram and one with the mask applied to the word following the bigram. This is the method used in the main paper.
        
        We conduct a small-scale experiment on ``The Affordable Care Act'' topic to evaluate the performance of each of the aforementioned methods. The experiment is conducted as described in Section  \ref{sct:methodology} but only on one topic. As is shown in Table \ref{tab:small-mask}, the Bigram outer (BO) achieves the best performance among all methods on ``SoA-s'', ``SoA-t'', and ``human''.Therefore, we utilize it for our main experiments.
    
        \begin{table}[!hbt]\small
            \centering
            \begin{tabular}{l|llll}
            \hline
                 & SoA-t         & SoA-s         & MBR      &human     \\ \hline
            RS   & 0.29          & 0.26          & 0.26     &0.48     \\
            BERT & 0.28          & 0.23          & 0.25     & \textbf{0.60}        \\
            BI   & 0.28          & 0.24          & 0.25     &0.50     \\
            TI   & 0.28          & 0.24          & 0.29     &\textbf{0.60}     \\
            BO   & \textbf{0.32} & \textbf{0.30} & \textbf{0.30} &0.56 \\ \hline
            \end{tabular}
            \caption{Comparison of agreement ($\tau$) between ground truth data and media bias estimations generated using different mask and prompt selection methods. Agreements are calculated using Kendall's $\tau$ for the ``The Affordable Care Act'' topic.}
            \label{tab:small-mask}
        \end{table}    
    
    \subsection{Masked Prompts Statistics}
    \label{sct:bigram-outer-stats}
    In Table \ref{tab:masked_prompts}, we show the number of bigrams, instances in the dev set, and masked prompts (note that each instance produces one or more masked prompts) for each topic for the bigram outer method. We observe that other than the ``The Affordable Care Act'' all topics have a similar amount of data.
    
    \label{sct:bigrams-and-instances}
        \begin{table*}[!hbt]
            \centering
            \begin{tabular}{lllllll}
            \hline
                                 & Climate & Cor & Drug & SSM & Care & Total  \\
            \hline
            \# of bigrams        & 39             & 37            & 37          & 31           & 75                      & 219    \\
            \# of instances      & 1,856          & 2,067         & 1,992       & 1,514        & 4,071                   & 11,500 \\
            \# of masked prompts & 4,299          & 5,152         & 5,161       & 3,550        & 13,160                  & 31,322 \\
            \hline
            \end{tabular}
            \caption{The statistics of the number of bigrams, instances in the dev set, and masked prompts for the BO prompt and mask selection method. ``Climate'' : ``Climate Change'', ``Cor'' : ``Corporate Tax'', ``Drug'' : ``Drug policy'', ``SSM'' : ``Gay Marriage'', ``Care'' : ``The Affordable Care Act''.}
            \label{tab:masked_prompts}
        \end{table*}

\section{Experiments}
\setcounter{table}{0}
    \label{sct:expe}
    \subsection{Computing Infrastructure}
    In our experiments, we utilized a Lambda machine with 250 GB of memory, 4 RTX 8000 GPUs, and 16 CPU cores. The operation system of the machine is Ubuntu 16.04. Our experiments are conducted with Python 3.8.8 with the following packages: gensim (4.0.1), grakel (0.1.8), joblib (1.0.1), matpolitlib (3.3.4), nltk (3.6.1), numpy (1.20.1), scikit-learn (0.24.1), scipy (1.6.2), torch (1.9.0), transformers (4.8.2), tqdm (4.69.0), zss (1.2.0). The CUDA version is 11.2 and the GPU Driver Version is 460.67
    \subsection{Hyperparameters and Random Seed}
        \label{sct:hyper-and-seed}
        In our experiments, all random seeds are set to be 42.
        We utilize the ``Hugging Face'' implementation for fine-tuning the language model. During the fine-tuning, because of the limits of the GPU memory, we set the batch size as 16. The training epoch is set to be 10. All the other hyper parameters for the training process are set to be the default value of the package.

        For training the LDA baseline, we utilize the ``LdaMulticore'' from ``gensim''. When building the word dictionary of the LDA baseline, we keep the most frequent 2,000,000 words, which is the default hyper-parameters of the ``Dictionary'' library of ``gensim''.  For the ``LdaMulticore'', we set the number of topics to be 10, random seed as 42, ``workers'' as 4, ``passes'' to be 2, and keep all the other hyper-parameters as the default value.
        
        For training the Tf-IDF baseline, we utilize the ``TfidfVectorizer'' of ``sklearn''. We set the ``ngram\_range'' to be ``(1,3)'' which means the TF-IDF model will consider unigram, bigram, and trigram tokens. We limit to bigrams because it achieved the best performance. We keep all the other parameters to be the default value.
        
        For the LM-c baseline, we utilize a language model for the task of predicting the source of instances. We set the epoch to be 10 and test the performance at the end of each epoch. We use the model that achieves the best performance for encoding the instances in the dev sets.
        
        For the LM-m baseline, we utilize the fine-tuned model without any change.
        
        When conducting hierarchical clustering (see Section \ref{sct:cluster}), we utilize the ``AgglomerativeClustering'' library from  ``sklearn''. We set the ``linkage'' to be ``complete'' and keep all the default hyper-parameters.

    \subsection{Steps for Reproducing our Results}
    \label{sec:repr}
        As part of the supplementary material, we have included both the code and data for reproducing our results. After downloading the ``code'' and ``data'', please unzip them and put all folders from the ``data'' folder in the ``code'' folder. To reproduce the work, for each topic (``climate-change'', ``corporate-tax'', ``drug-policy'', ``gay-marriage'' and ``obamacare''), please run the following three shells in sequence: ``./shell/data\_collect.sh'', ``./shell/train\_lm.py'' and ``./shell/media\_encode.sh''. All these three shells accept the name of the topic as the parameter.
        
        To calculate the performance of baselines, pleases run the following three shells:``./shell/run\_mlm.sh'', ``./shell/run\_class.sh'', and ``./shell/get\_baseline.sh''

\section{Human Evaluation}
    \subsection{Dataset Construction}
    \label{sct:human-eval-data}
    We rely on human annotation data for evaluating our framework. We recruit five students from the United States (3 undergrads and 2 doctoral; 3 female and 2 male) with knowledge of the U.S. political landscape. We choose five of the ten media outlets (Breitbart News Network, CBS News, CNN, Fox News, and Wall Street Journal). We first provide students with 75 sample articles (3 articles on each of the 5 topics for each outlet) for reference. Next, we ask the respondents to evaluate the similarity among the 5 media outlets. Specifically, for each media outlet pair (e.g. CNN and Fox News), we ask the respondents to choose which of the two outlets they consider ideologically closest to each of the remaining three outlets. We also asked the respondents to rate their confidence in each choice. 
    
\begin{figure}[!hbt]
        \centering
        \includegraphics[width=1.00\columnwidth]{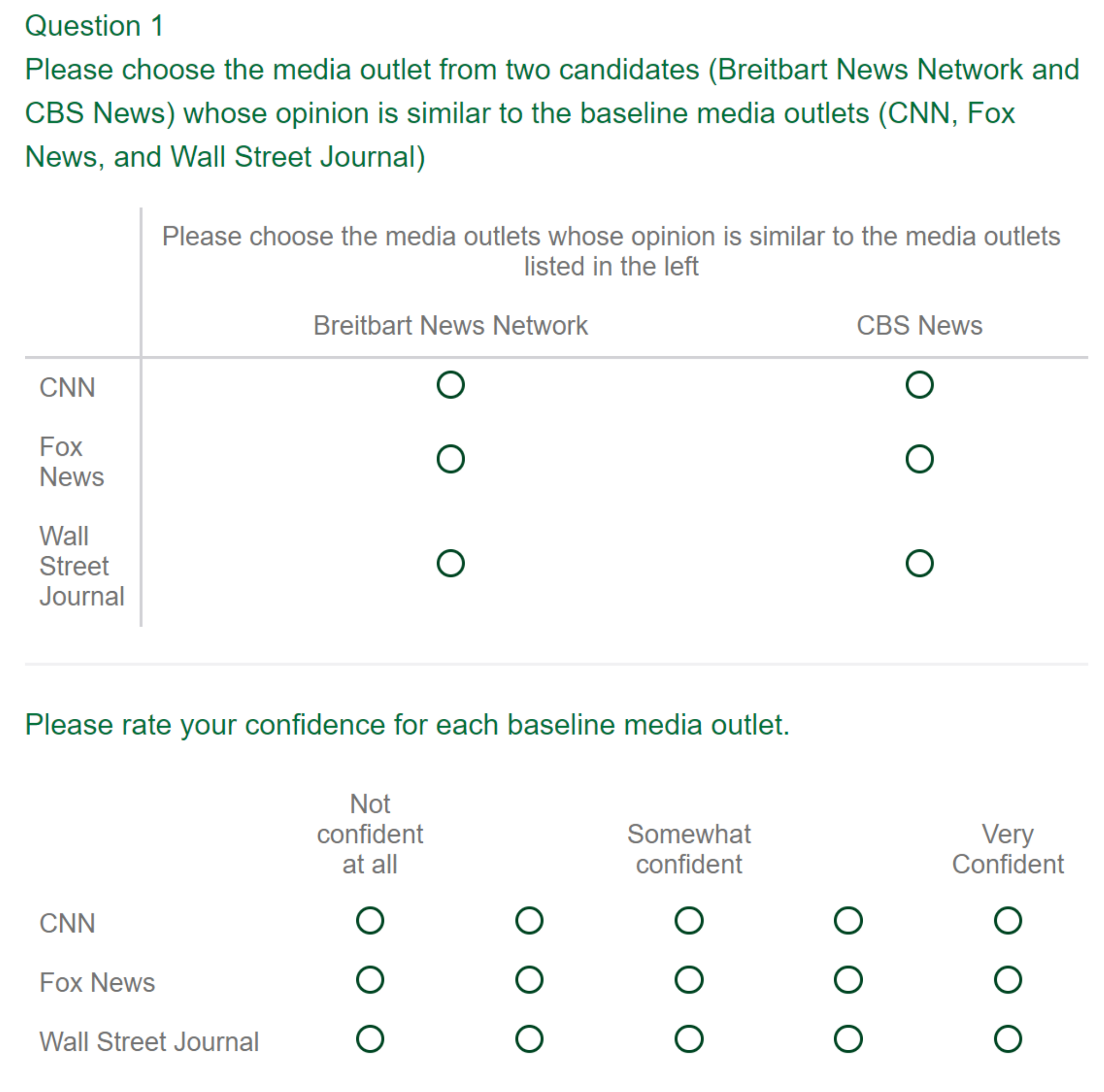}
        \caption{Example of the human evaluation}
        \label{fig:survey}
    \end{figure}

    An example of the survey question is shown in Figure \ref{fig:survey}. In this example, respondents are required to choose whether Breitbart News Network or CBS News is more similar to CNN, Fox News, and Wall Street Journal, along with the confidence for their selection.
    
    \subsection{Results}
    \label{sct:human-eval-stats}
In Table \ref{tab:main-performance-details}, we show the performance of each combination of prompt generation method, normalization method, and language models. We can observe that except for the choice of LM, the performance of our model is dependent on these choices.
\begin{table*}[!hbt]\small
    \begin{tabular}{l|l|ll|ll|ll|l}
    \hline
                                  &         & \multicolumn{2}{l|}{dealarative}& \multicolumn{2}{l|}{interrogative}& \multicolumn{2}{l|}{association}& bigram outer \\ \hline
                                  &         & q\&a           & single         & q\&a            & single          & q\&a           & single         &              \\ \hline
    \multirow{3}{*}{RoBERTa}      & no     & 0.28(0.27)     & 0.45(0.09)     & 0.07(0.23)      & 0.46(0.1)       & 0.41(0.21)     & 0.28(0.09)     & 0.45(0.09)   \\
                                  & general & 0.38(0.10)     & 0.41(0.14)     & 0.44(0.17)      & 0.45(0.17)      & 0.41(0.13)     & 0.44(0.14)     & 0.49(0.16)   \\
                                  & domain  & 0.42(0.11)     & 0.39(0.15)     & 0.45(0.13)      & 0.46(0.20)      & 0.38(0.10)     & 0.38(0.18)     & 0.50(0.07)   \\ \hline
    \multirow{3}{*}{BERT-un}      & no     & 0.24(0.00)     & 0.54(0.16)     & 0.26(0.10)      & 0.33(0.15)      & 0.24(0.00)     & 0.26(0.04)     & 0.41(0.13)   \\
                                  & general & 0.35(0.06)     & 0.45(0.19)     & 0.48(0.15)      & 0.37(0.16)      & 0.30(0.12)     & 0.45(0.33)     & 0.36(0.11)   \\
                                  & domain  & 0.30(0.13)     & 0.50(0.18)     & 0.46(0.15)      & 0.42(0.27)      & 0.38(0.15)     & 0.41(0.32)     & 0.48(0.08)   \\ \hline
    \multirow{3}{*}{BERT-ca}      & no     & 0.27(0.04)     & 0.34(0.11)     & 0.23(0.14)      & 0.43(0.22)      & 0.26(0.04)     & 0.24(0.00)     & 0.34(0.08)   \\
                                  & general & 0.34(0.17)     & 0.49(0.11)     & 0.35(0.09)      & 0.49(0.13)      & 0.33(0.12)     & 0.54(0.06)     & 0.42(0.11)   \\
                                  & domain  & 0.45(0.22)     & 0.53(0.08)     & 0.43(0.23)      & 0.58(0.16)      & 0.38(0.17)     & 0.52(0.11)     & 0.47(0.10)   \\  \hline
    \end{tabular}
        \caption{Agreement between media similarity rankings using our framework and the human evaluation dataset. Agreement is calculated using Kendall's $\tau$ (higher magnitude is better). The results show the average performance across 5 topics. The standard deviations are shown in the brackets. ``RoBERTa'' : ``RoBERTa-base'', ``BERT-un'' : ``BERT-base-uncased'', ``BERT-ca'' : ``BERT-base-cased''.  ``no'': ``no-normalization'', ``general'' : ``general-normalization'' ,  ``domain'' : ``domain-normalizatio'', ``q\&a'' : question-answer pair pattern, ``single'' : single sentence pattern}
        \label{tab:main-performance-details}
\end{table*}

\section{Automatic Evaluation}
\setcounter{table}{0}
    \label{sct:auto-eval}
    Except for the human evaluation, we also evaluate the performance of our framework with three independent ground-truth datasets based on survey data from Pew Research \cite{pew2020news} and Allsides.com.
    
    \subsection{Ground Truth Dataset Construction}
    \label{sct:data-g}
    We construct three ground truth datasets for automatic evaluation purpose. Two datasets, SoA-s and SoA-t, are based on survey data from Pew Research \cite{pew2020news}. The raw data of the SoA-s (source) dataset includes the share of all US adults that read news articles from each outlet, while the SoA-t (trust) dataset includes the share of all US adults who trust these news outlets. Both datasets include the share of all US adults regardless of ideology and all US adults with a particular ideology.
    To reduce the influence of the popularity of particular outlets, we normalize the data of each ideology by the share of all US adults for each outlet. For example, 47\% of all U.S adults trust CNN, and 67\% of Democrats trust CNN. The normalized score for CNN for Democrat will thus be 1.43. Table \ref{tab:SoA-s example} two examples for the SoA-s dataset. There are 7 different labels showing different parties and ideologies. The ``ALL U.S. adults '' is used as the baseline for measuring the tendency of trusting media outlets for people with different political leanings.
    
    \begin{table}[!hbt]\scriptsize
        \begin{tabular}{l|rrrr}
                                                                                      & \multicolumn{2}{l}{Original}                                 & \multicolumn{2}{l}{Processed}                                \\
                                                                                      & \multicolumn{1}{l}{ABC News} & \multicolumn{1}{l}{Breitbart} & \multicolumn{1}{l}{ABC News} & \multicolumn{1}{l}{Breitbart} \\
        All U.S. adults                                                               & 33\%                         & 4\%                            & 1.00                         & 1.00                          \\\\ \hline
        \begin{tabular}[c]{@{}l@{}}Democrat/\\ Lean Dem\end{tabular}                  & 37\%                         & 0\%                            & 1.12                         & 0.00                          \\\\ \hline
        \begin{tabular}[c]{@{}l@{}}Republican/\\ Lean Rep\end{tabular}                & 30\%                         & 8\%                            & 0.91                         & 2.00                          \\\\ \hline
        \begin{tabular}[c]{@{}l@{}}Liberal Dem/\\ Lean Dem\end{tabular}               & 32\%                         & 0\%                            & 0.97                         & 0.00                          \\\\ \hline
        \begin{tabular}[c]{@{}l@{}}Conservative/\\ Moderate Dem/\\Lean Dem\end{tabular} & 42\%                         & 1\%                            & 1.27                         & 0.25                          \\\\ \hline
        \begin{tabular}[c]{@{}l@{}}Moderate/\\ Liberal Rep/\\Lean Rep\end{tabular}      & 36\%                         & 3\%                            & 1.09                         & 0.75                          \\\\ \hline
        \begin{tabular}[c]{@{}l@{}}Conservative Rep/\\ Lean Rep\end{tabular}          & 26\%                         & 11\%                           & 0.79                         & 2.75                              \\\\ \hline
        \end{tabular}
            \caption{Examples from the SoA-s dataset. We show the original and processed data for ABC News and Breitbart.}
            \label{tab:SoA-s example}
    \end{table}

    The third dataset (MBR) is based on expert curation of media partisanship by Allsides.com \footnote{https://www.allsides.com/media-bias/media-bias-ratings}. This dataset labels each media outlet with five political leanings, from left to right, based on editorial review, third-party analysis, independent review, surveys, and community feedback.
    
    We measure the relative rankings of the outlets using these ground truth datasets using a similar procedure as described in Section \ref{mrb}: for each outlet, we calculate the distance to the other outlets (based on their ground truth political ideology labels) and create a ranked list. For SoA-s and SoA-t we use cosine distance (since these datasets provide a distribution over ideologies for each outlet), while for MBR we look at the absolute distance between the outlets (since this dataset provides a single ideological score for each outlet).
    
    \subsection{Results}
    \label{sct::auto-results}
    
    The Kendall's $\tau$ between different ground truth datasets (SoA-t and MBR) is only about 0.5. We believe this can be seen as an upper bound for our experiments.
    
    Table \ref{tab:auto-pattern-stats} shows the effect of prompt generation methods on this task. We can observe that the single-sentence pattern achieves better performance than the question-answer pair patterns. Different than the human evaluations, we see that the performance of the automatic prompt generation method bigram outer is better than the manual methods. This is especially true for the declarative manual method. This could be due to the limited number of manual prompts created for these experiments.
    \begin{table*}[hbt]
    \centering
    \begin{tabular}{lllll}
    \hline
                                   &        & SoA-s      & SoA-t      & MBR        \\ \hline
    \multirow{2}{*}{declarative}   & q\&a   & 0.09(0.12) & 0.11(0.10) & 0.12(0.09) \\
                                   & single & 0.20(0.03) & 0.20(0.04) & 0.19(0.04) \\ \hline
    \multirow{2}{*}{interrogative} & q\&a   & 0.12(0.12) & 0.14(0.10) & 0.12(0.09) \\
                                   & single & 0.24(0.05) & 0.23(0.04) & 0.18(0.03) \\ \hline
    \multirow{2}{*}{association}   & q\&a   & 0.12(0.12) & 0.14(0.09) & 0.14(0.08) \\
                                   & single & 0.15(0.13) & 0.16(0.11) & 0.16(0.10) \\ \hline
    bigram outer                   &        & 0.27(0.04) & 0.27(0.05) & 0.23(0.04) \\ \hline
    \end{tabular}
    \caption{Agreement ($\tau$) between media similarity rankings using our framework and the three ground truth datasets. ``q\&a'' : question-answer pair pattern, ``single'' : single sentence pattern.}
    \label{tab:auto-pattern-stats}
    \end{table*}

    Table \ref{tab:auto-norm-stats} shows the influence of normalization methods on this task. We can observe that two normalization methods still achieve better performance than the no-normalization method. When focusing on the ``no-normalization'' method, we can observe that it's only slightly better than a random guess which suggests that it's not a good choice for this task.
    \begin{table}[!hbt]
    \centering
    \begin{tabular}{llll}
    \hline
            & SoA-s      & SoA-t      & MBR        \\ \hline
    no      & 0.06(0.14) & 0.08(0.12) & 0.07(0.09) \\
    general & 0.22(0.12) & 0.21(0.10) & 0.21(0.08) \\
    domain  & 0.22(0.11) & 0.22(0.09) & 0.19(0.08) \\ \hline
    \end{tabular}
    \caption{Agreement ($\tau$) between media similarity rankings using our framework and three ground truth datasets. ``no'' : ``no-normalization'', ``general'' : ``general-normalization'', ``domain'' : ``domain-normalization''.}
    \label{tab:auto-norm-stats}
    \end{table}
    
    Table \ref{tab:auto-lm-stats} shows the influence of different LMs in our framework. We see that our model is robust to the choice of LMs, with slight differences between the models.
    \begin{table}[hbt]
    \centering
    \begin{tabular}{llll}
    \hline
            & SoA-s      & SoA-t      & MBR        \\ \hline
    RoBERTa & 0.18(0.11) & 0.18(0.10) & 0.17(0.07) \\
    BERT-un & 0.17(0.11) & 0.18(0.09) & 0.17(0.08) \\
    BERT-ca & 0.17(0.11) & 0.18(0.09) & 0.15(0.08) \\ \hline
    \end{tabular}
    \caption{Agreement ($\tau$) between media similarity rankings using our framework and three ground truth datasets. ``RoBERTa'' : ``RoBERTa-base'', ``BERT-un'' :``BERT-base-uncased'', ``BERT-ca'' : ``BERT-base-cased''.}
    \label{tab:auto-lm-stats}
    \end{table}    

    In Table \ref{tab:side-performance}, we show the performance of all combinations of our framework on the ground truth datasets SoA-s, SoA-t, and MBR. Our results show that the three factors of our framework (prompt generation, normalization, and choice of LM) should be set together. 
    
    \begin{table*}[!hbt]\scriptsize
    \centering
    \begin{tabular}{ll|l|ll|ll|ll|l}
    \hline
                           &                               &         & \multicolumn{2}{l}{declarative} & \multicolumn{2}{l}{interrogative} & \multicolumn{2}{l}{association} & bigram outer \\ \hline
                           &                               &         & q\&a            & single        & q\&a             & single         & q\&a           & single         &              \\ \hline
    \multirow{9}{*}{SoA-s} & \multirow{3}{*}{RoBERTa}      & no     & -0.08(0.11)     & 0.13(0.12)    & -0.04(0.23)      & 0.23(0.11)     & 0.07(0.2)      & -0.02(0.05)    & 0.29(0.14)   \\
                           &                               & general & 0.16(0.09)      & 0.20(0.09)     & 0.18(0.08)       & 0.32(0.06)     & 0.23(0.11)     & 0.26(0.09)     & 0.17(0.06)   \\
                           &                               & domain  & 0.18(0.07)      & 0.22(0.09)    & 0.17(0.1)        & 0.31(0.07)     & 0.24(0.05)     & 0.27(0.07)     & 0.25(0.11)   \\ \hline
                           & \multirow{3}{*}{BERT-un}      & no     & -0.05(0.01)     & 0.20(0.16)     & -0.05(0.06)      & 0.26(0.13)     & -0.05(0.01)    & -0.0(0.11)     & 0.3(0.07)    \\
                           &                               & general & 0.16(0.02)      & 0.23(0.15)    & 0.15(0.05)       & 0.21(0.10)      & 0.15(0.06)     & 0.26(0.10)      & 0.31(0.07)   \\
                           &                               & domain  & 0.17(0.05)      & 0.20(0.11)     & 0.18(0.08)       & 0.2(0.11)      & 0.17(0.08)     & 0.22(0.08)     & 0.27(0.11)   \\ \hline
                           & \multirow{3}{*}{BERT-ca}      & no     & -0.06(0.01)     & 0.23(0.04)    & -0.00(0.08)       & 0.2(0.06)      & -0.07(0.01)    & -0.05(0.00)     & 0.27(0.07)   \\
                           &                               & general & 0.20(0.10)        & 0.24(0.12)    & 0.23(0.06)       & 0.22(0.06)     & 0.15(0.07)     & 0.2(0.05)      & 0.32(0.09)   \\
                           &                               & domain  & 0.17(0.02)      & 0.18(0.07)    & 0.24(0.05)       & 0.18(0.06)     & 0.21(0.06)     & 0.21(0.04)     & 0.28(0.10)    \\ \hline \\ \hline
    \multirow{9}{*}{SoA-t} & \multirow{3}{*}{RoBERTa}      & no     & -0.04(0.11)     & 0.14(0.12)    & -0.02(0.21)      & 0.2(0.07)      & 0.1(0.19)      & 0.03(0.05)     & 0.29(0.11)   \\
                           &                               & general & 0.16(0.09)      & 0.19(0.10)     & 0.19(0.08)       & 0.3(0.06)      & 0.22(0.12)     & 0.25(0.08)     & 0.15(0.06)   \\
                           &                               & domain  & 0.18(0.10)       & 0.22(0.11)    & 0.19(0.09)       & 0.29(0.06)     & 0.23(0.04)     & 0.25(0.06)     & 0.27(0.08)   \\ \hline
                           & \multirow{3}{*}{BERT-un}      & no     & -0.00(0.00)       & 0.17(0.16)    & 0.00(0.07)        & 0.26(0.12)     & -0.00(0.00)      & 0.04(0.08)     & 0.3(0.07)    \\
                           &                               & general & 0.16(0.06)      & 0.25(0.14)    & 0.16(0.06)       & 0.22(0.08)     & 0.16(0.03)     & 0.26(0.10)      & 0.28(0.08)   \\
                           &                               & domain  & 0.16(0.05)      & 0.21(0.10)     & 0.17(0.06)       & 0.22(0.10)      & 0.18(0.07)     & 0.22(0.09)     & 0.27(0.11)   \\ \hline
                           & \multirow{3}{*}{BERT-ca}      & no     & -0.00(0.01)      & 0.25(0.09)    & 0.05(0.10)        & 0.2(0.03)      & -0.01(0.02)    & 0.0(0.00)       & 0.27(0.06)   \\
                           &                               & general & 0.21(0.13)      & 0.21(0.11)    & 0.24(0.06)       & 0.21(0.04)     & 0.17(0.05)     & 0.19(0.05)     & 0.3(0.08)    \\
                           &                               & domain  & 0.20(0.03)       & 0.18(0.09)    & 0.26(0.07)       & 0.18(0.09)     & 0.22(0.07)     & 0.19(0.08)     & 0.29(0.10)    \\ \hline \\ \hline
    \multirow{9}{*}{MBR}   & \multirow{3}{*}{RoBERTa}      & no     & -0.01(0.11)     & 0.18(0.12)    & -0.01(0.22)      & 0.14(0.13)     & 0.14(0.14)     & 0.02(0.01)     & 0.21(0.12)   \\
                           &                               & general & 0.18(0.1)       & 0.16(0.10)     & 0.20(0.11)        & 0.19(0.10)      & 0.27(0.14)     & 0.23(0.20)      & 0.24(0.07)   \\
                           &                               & domain  & 0.19(0.06)      & 0.19(0.10)     & 0.21(0.15)       & 0.17(0.11)     & 0.2(0.06)      & 0.24(0.12)     & 0.19(0.05)   \\ \hline
                           & \multirow{3}{*}{BERT-un}      & no     & 0.01(0.00)      & 0.16(0.15)    & 0.02(0.04)        & 0.21(0.12)     &0.01(0.0)	    &0.06(0.07)	      &0.23(0.13)   \\
                           &                               & general &0.18(0.04)	   &0.28(0.09)	   &0.18(0.06)	      &0.21(0.08)	    &0.16(0.08)	   &0.24(0.07)	     &0.3(0.09)\\
                           &                               & domain  & 0.15(0.05)	&0.22(0.05)	&0.17(0.05)	&0.2(0.08)	&0.19(0.08)	&0.2(0.06)	&0.2(0.09)                                   \\ \hline
                           & \multirow{3}{*}{BERT-ca}      & no     & 0.01(0.01)      & 0.19(0.09)    & -0.00(0.1)        & 0.18(0.07)     & 0.01(0.02)     & 0.01(0.00)      & 0.25(0.07)   \\
                           &                               & general & 0.19(0.04)      & 0.18(0.08)    & 0.16(0.05)       & 0.15(0.06)     & 0.16(0.09)     & 0.22(0.06)     & 0.27(0.11)   \\
                           &                               & domain  & 0.18(0.08)      & 0.15(0.07)    & 0.18(0.11)       & 0.15(0.08)     & 0.15(0.08)     & 0.23(0.06)     & 0.18(0.05)   \\  \hline
    \end{tabular}
            \caption{Agreement between media similarity rankings using our framework and the ``SoA-s'', ``SoA-t'' and ``MBR'' datasets. Agreement is calculated using Kendall's $\tau$ (higher magnitude is better). The results show the average performance of 5 topics. The standard deviations are shown in the brackets. ``RoBERTa'' : ``RoBERTa-base'', ``BERT-un'' :``BERT-base-uncased'', ``BERT-ca'' : ``BERT-base-cased'', ``no'' : ``no-normalization'', ``general'' : ``general-normalization'', ``domain'' : ``domain-normalizatio'' .``q\&a'' : question-answer pair pattern,  ``single'' : single sentence pattern.}
            \label{tab:side-performance}
    \end{table*}

\section{Error Analysis}
    
\setcounter{figure}{0}
    \label{sct::error-analysis}
    In Figure \ref{fig:distribution_all}, we show the $\tau$ distributions for all topics and datasets.
    \begin{figure*}[!hbt]
        \centering
        \includegraphics[width=0.6\columnwidth]{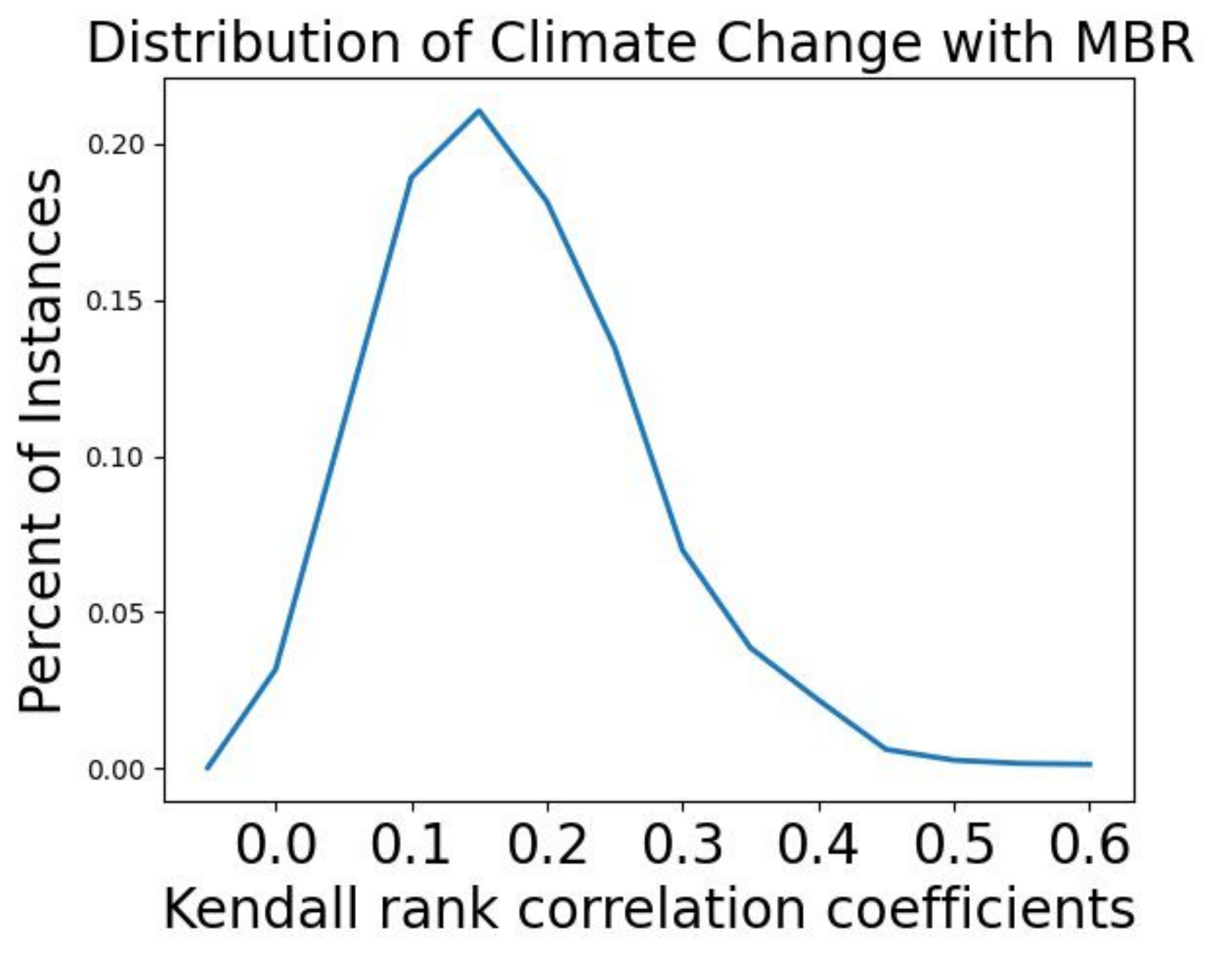}
        \includegraphics[width=0.6\columnwidth]{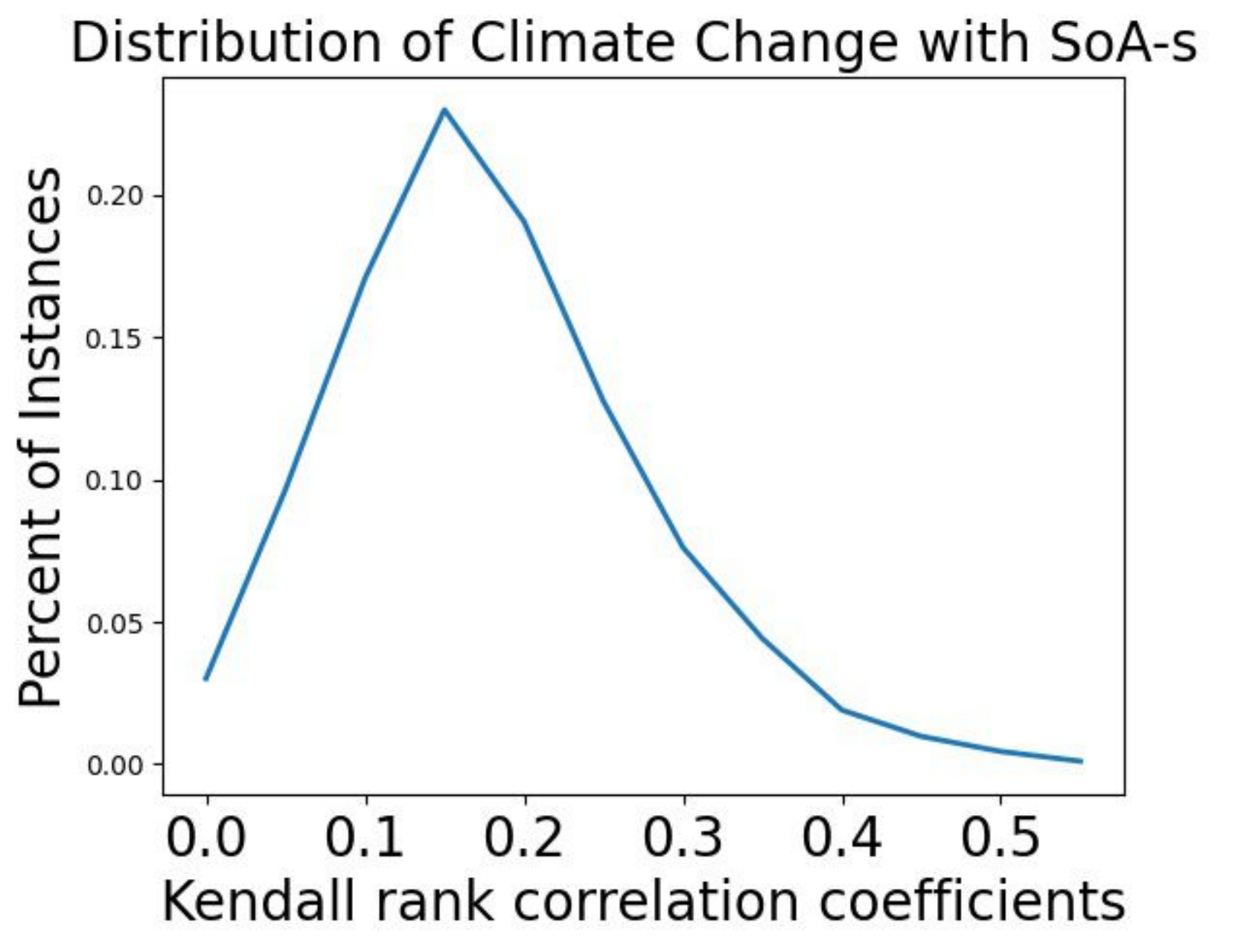}
        \includegraphics[width=0.6\columnwidth]{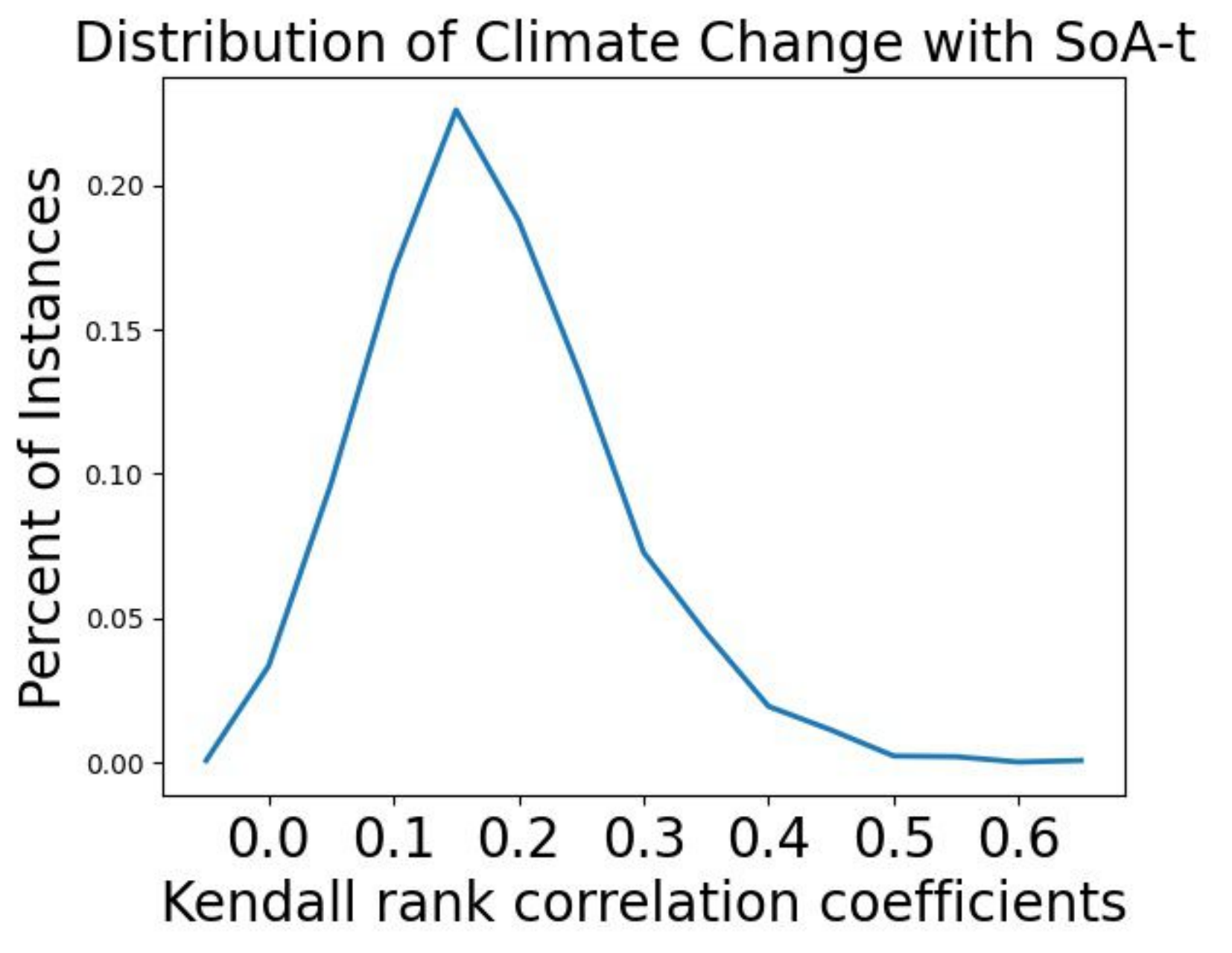}
        \includegraphics[width=0.6\columnwidth]{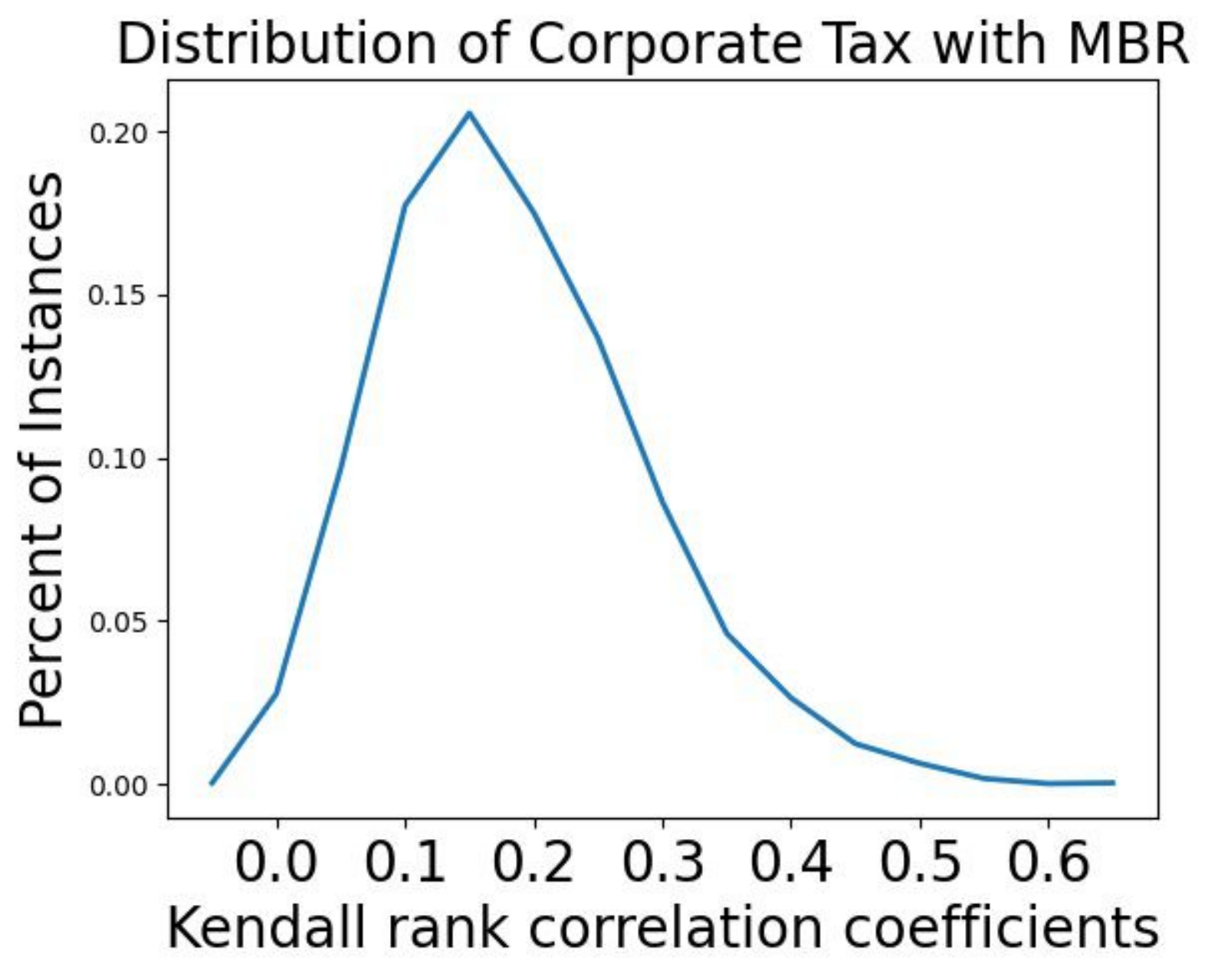}
        \includegraphics[width=0.6\columnwidth]{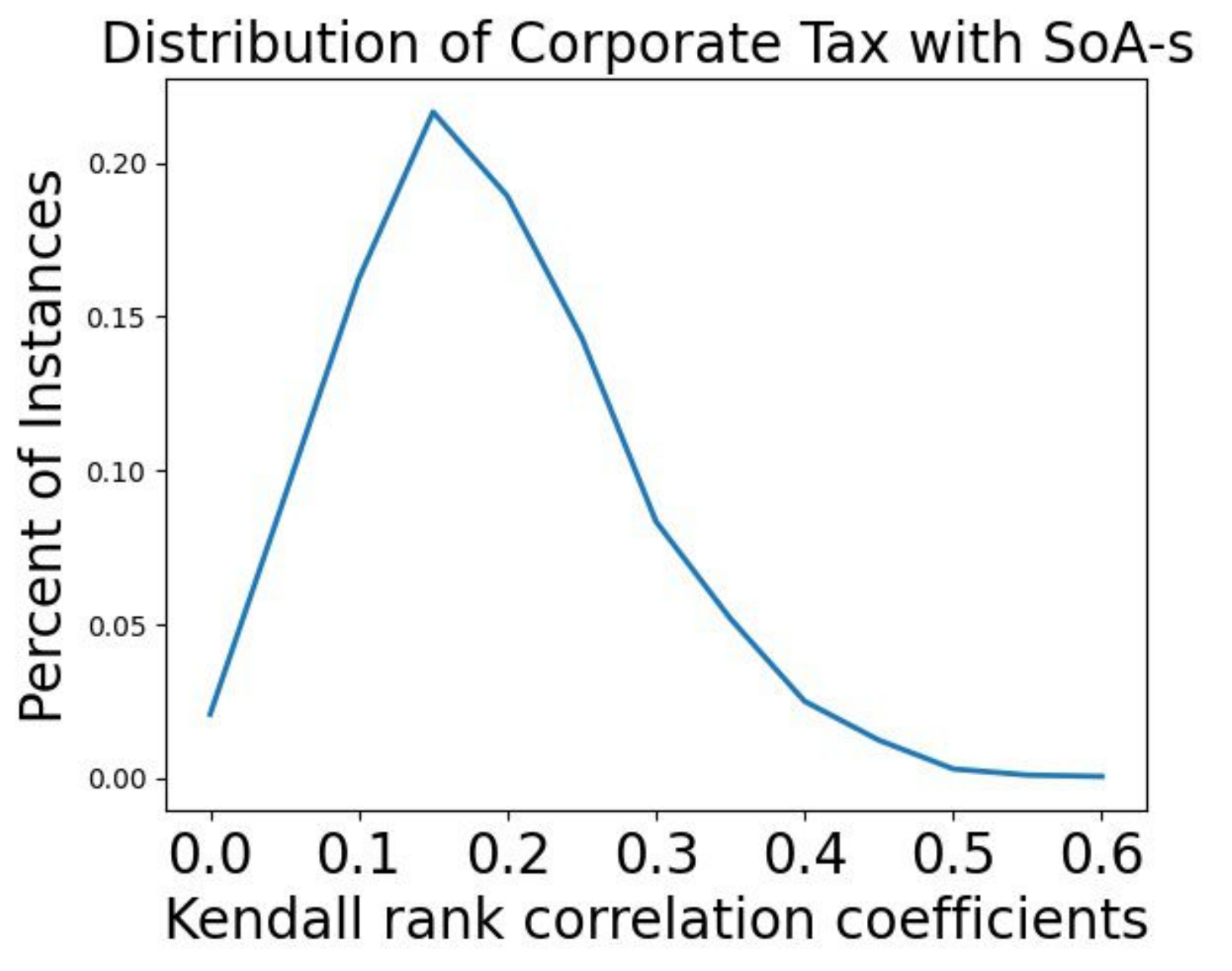}
        \includegraphics[width=0.6\columnwidth]{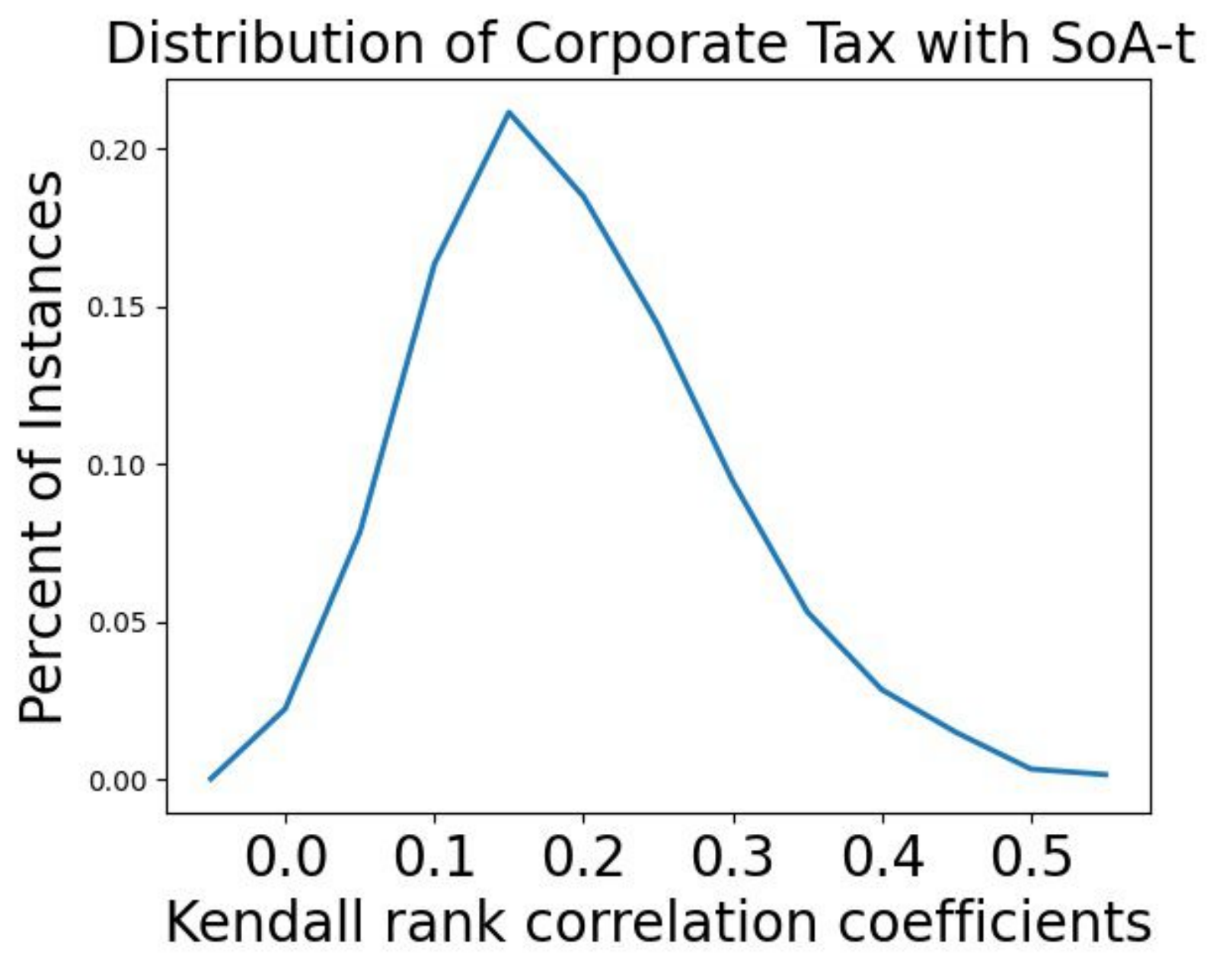}
        \includegraphics[width=0.6\columnwidth]{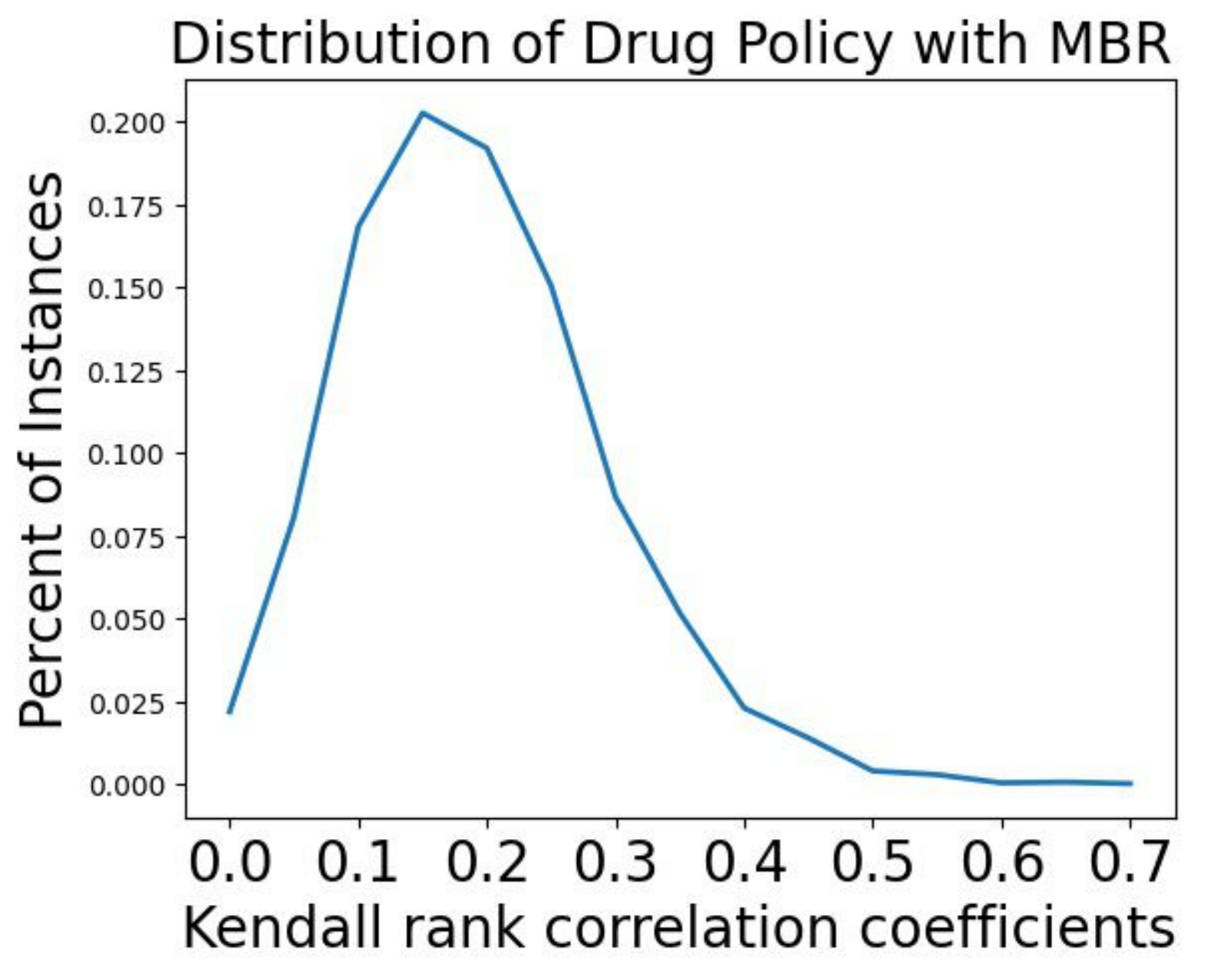}
        \includegraphics[width=0.6\columnwidth]{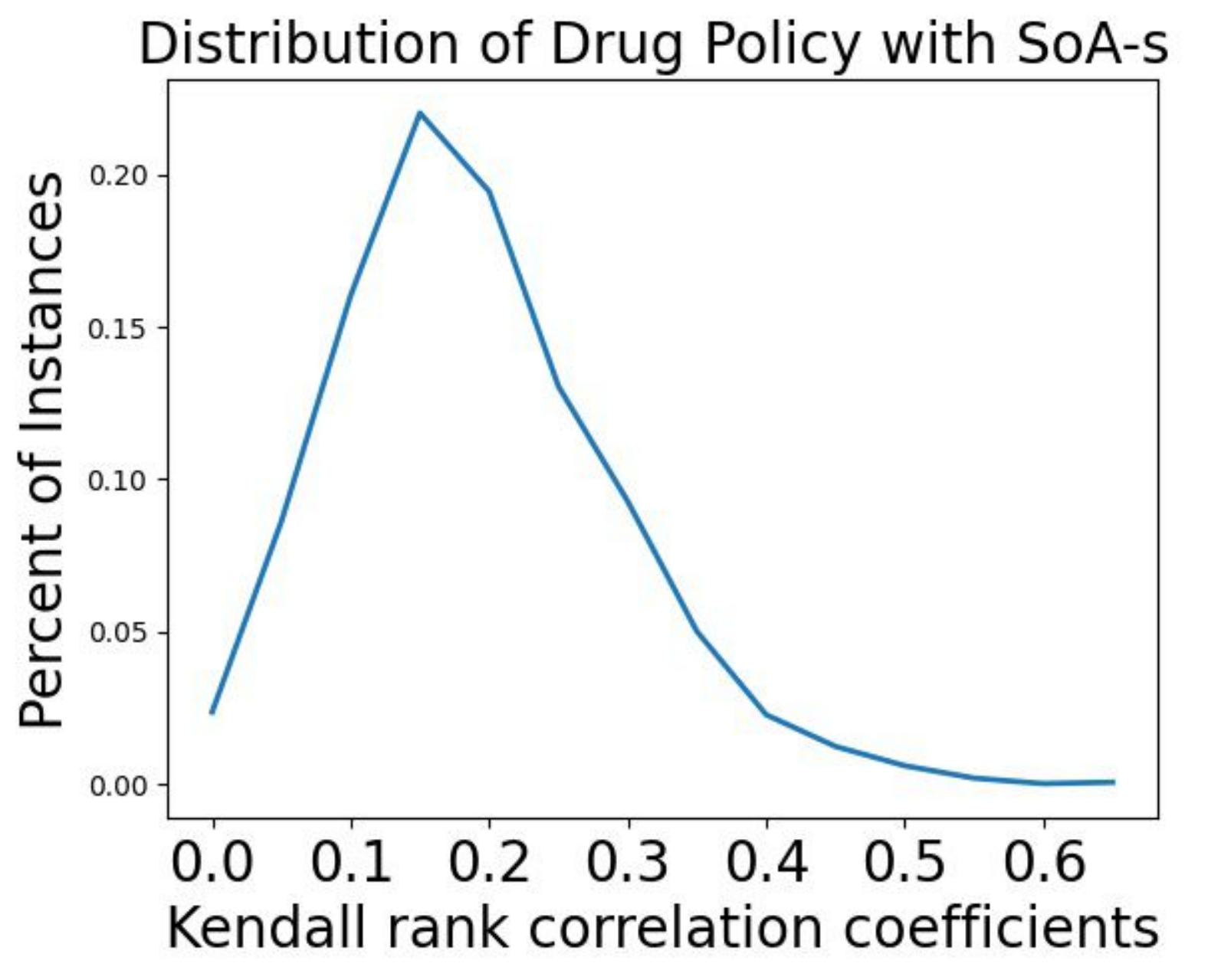}
        \includegraphics[width=0.6\columnwidth]{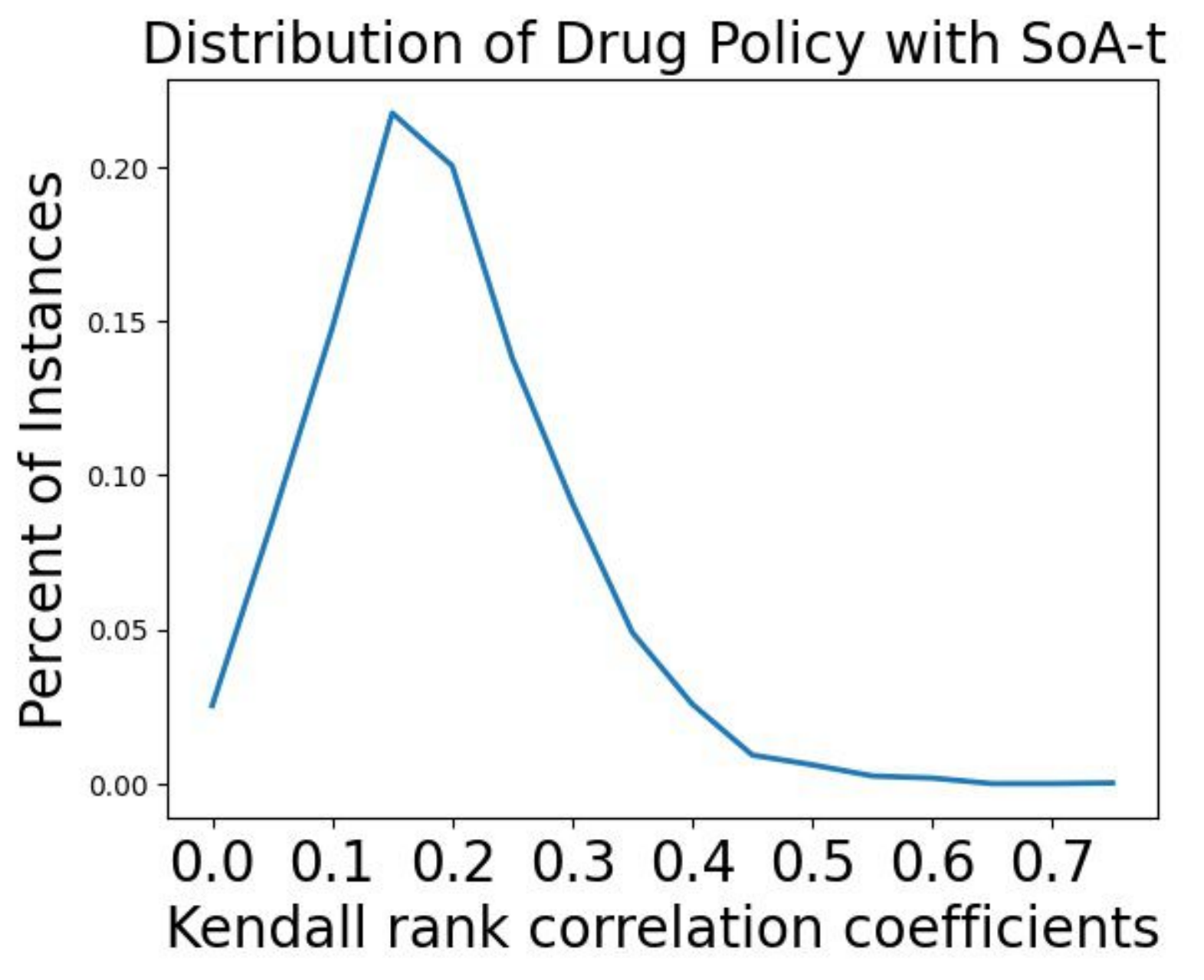}
        \includegraphics[width=0.6\columnwidth]{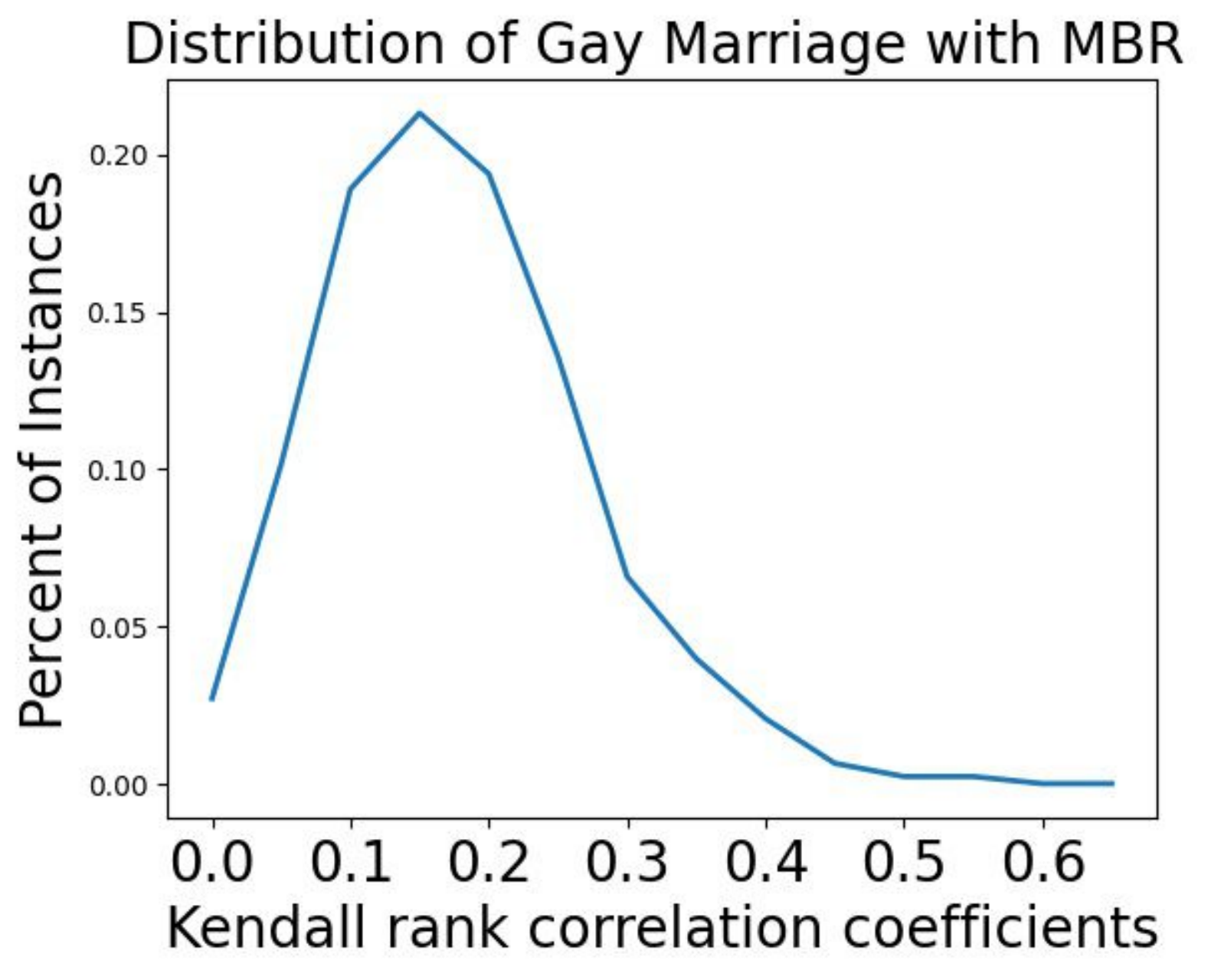}
        \includegraphics[width=0.6\columnwidth]{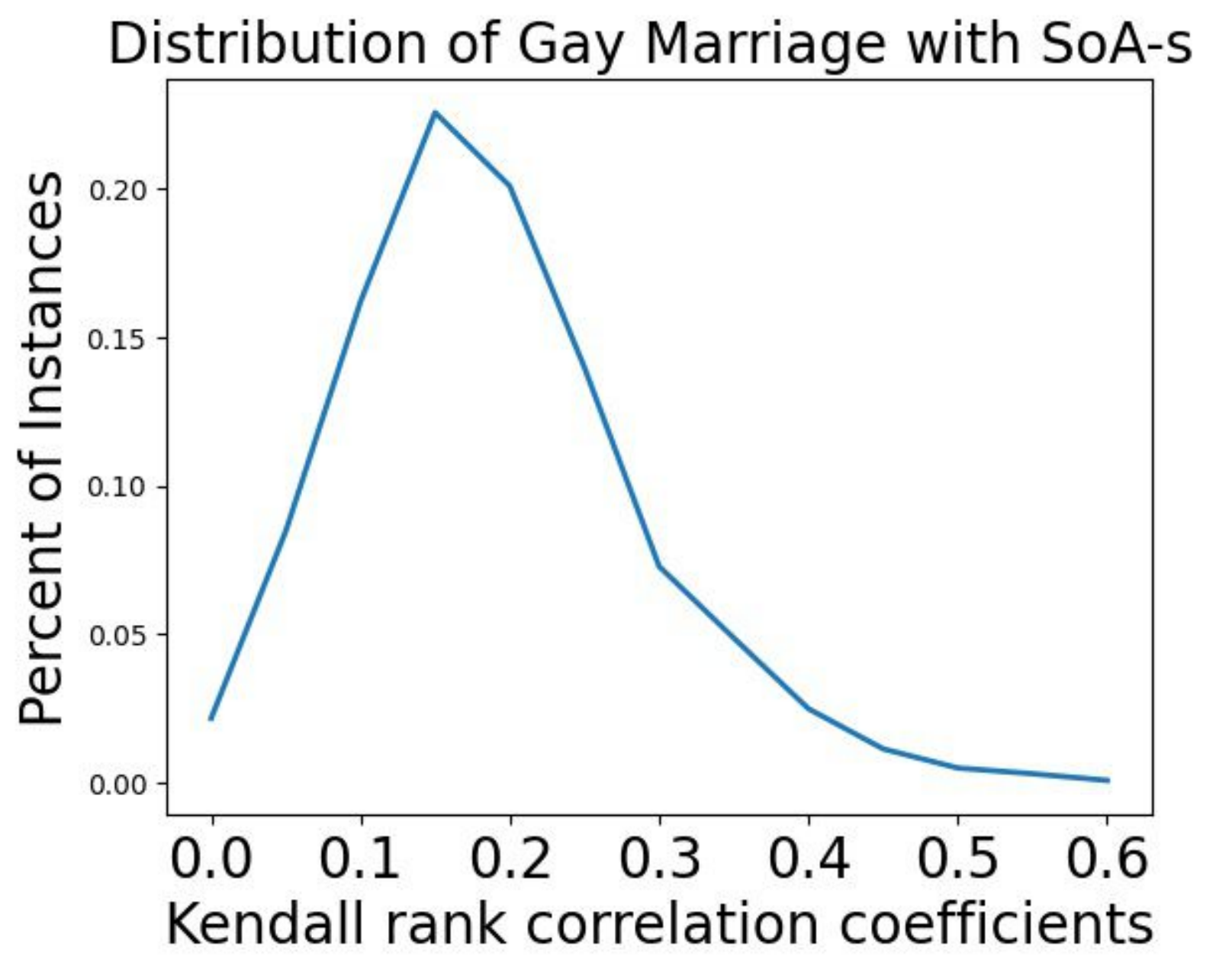}
        \includegraphics[width=0.6\columnwidth]{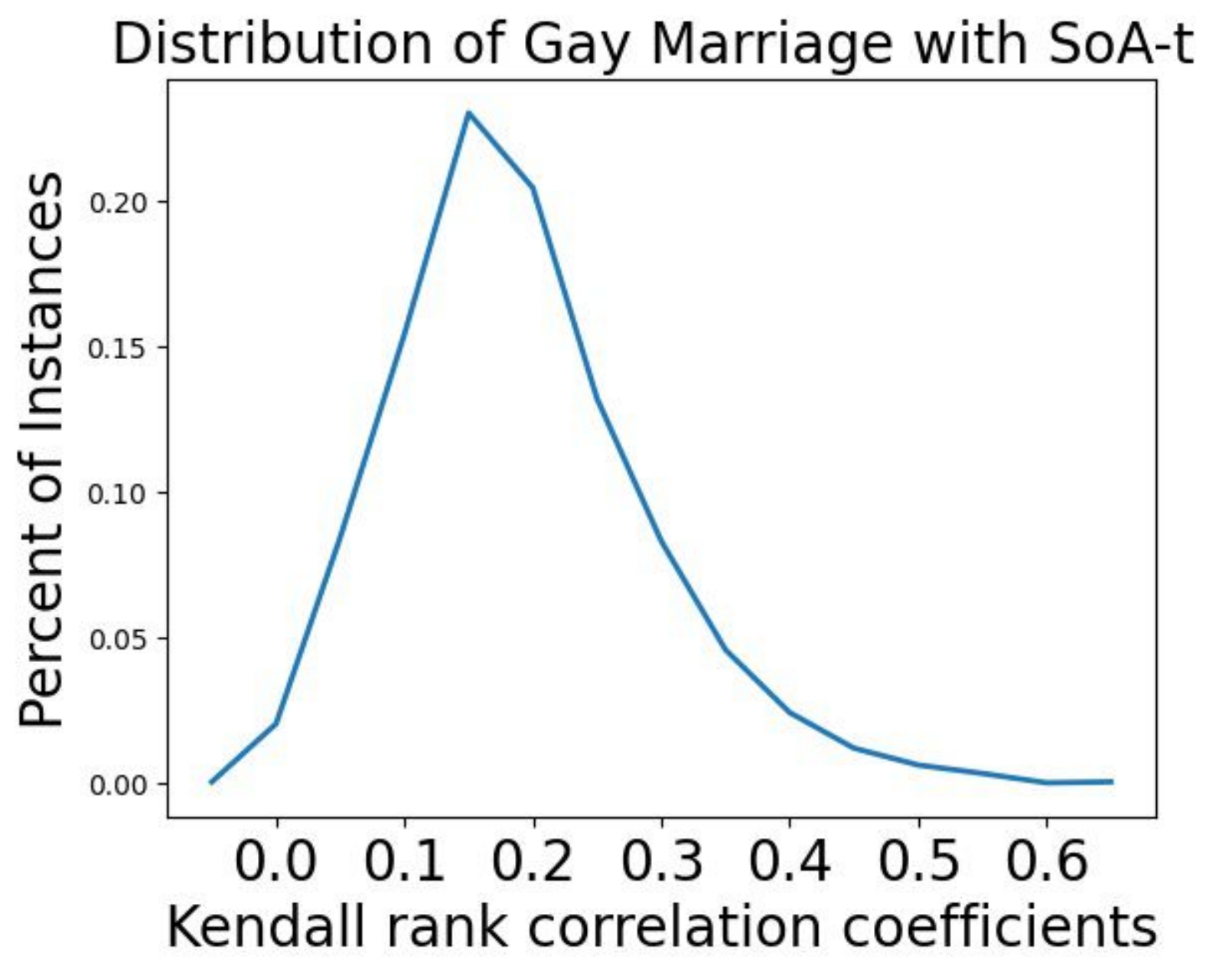}
        \includegraphics[width=0.6\columnwidth]{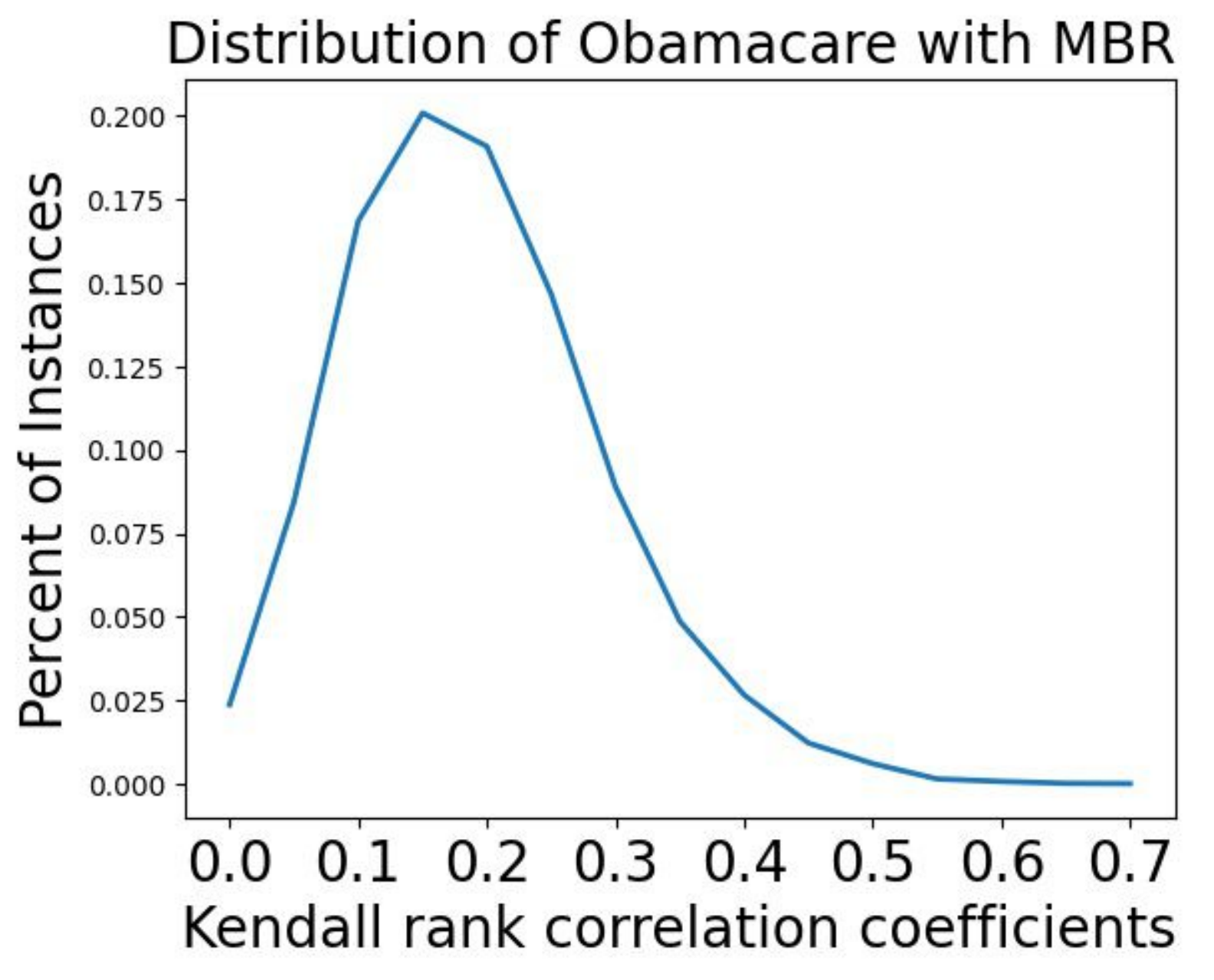}
        \includegraphics[width=0.6\columnwidth]{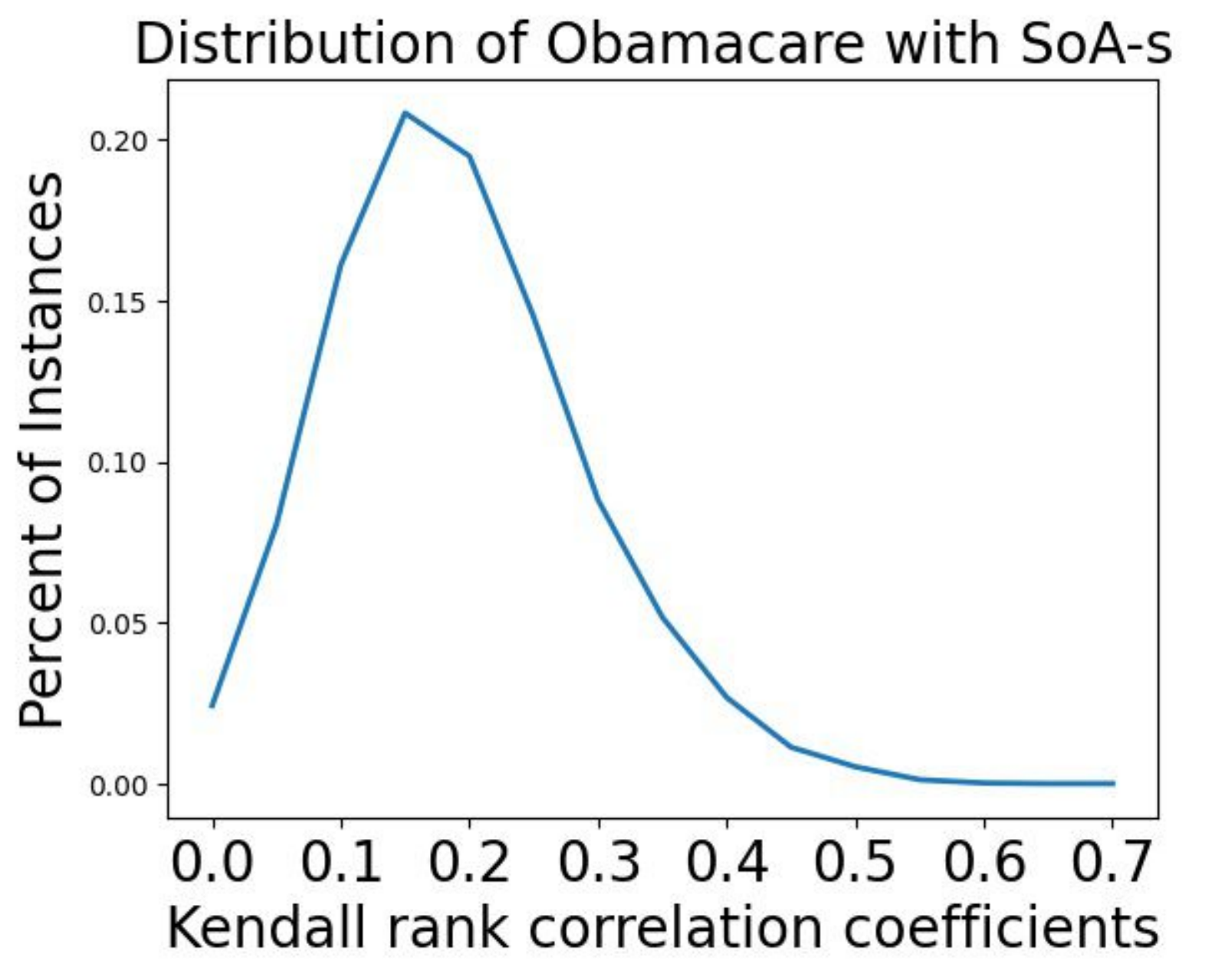}
        \includegraphics[width=0.6\columnwidth]{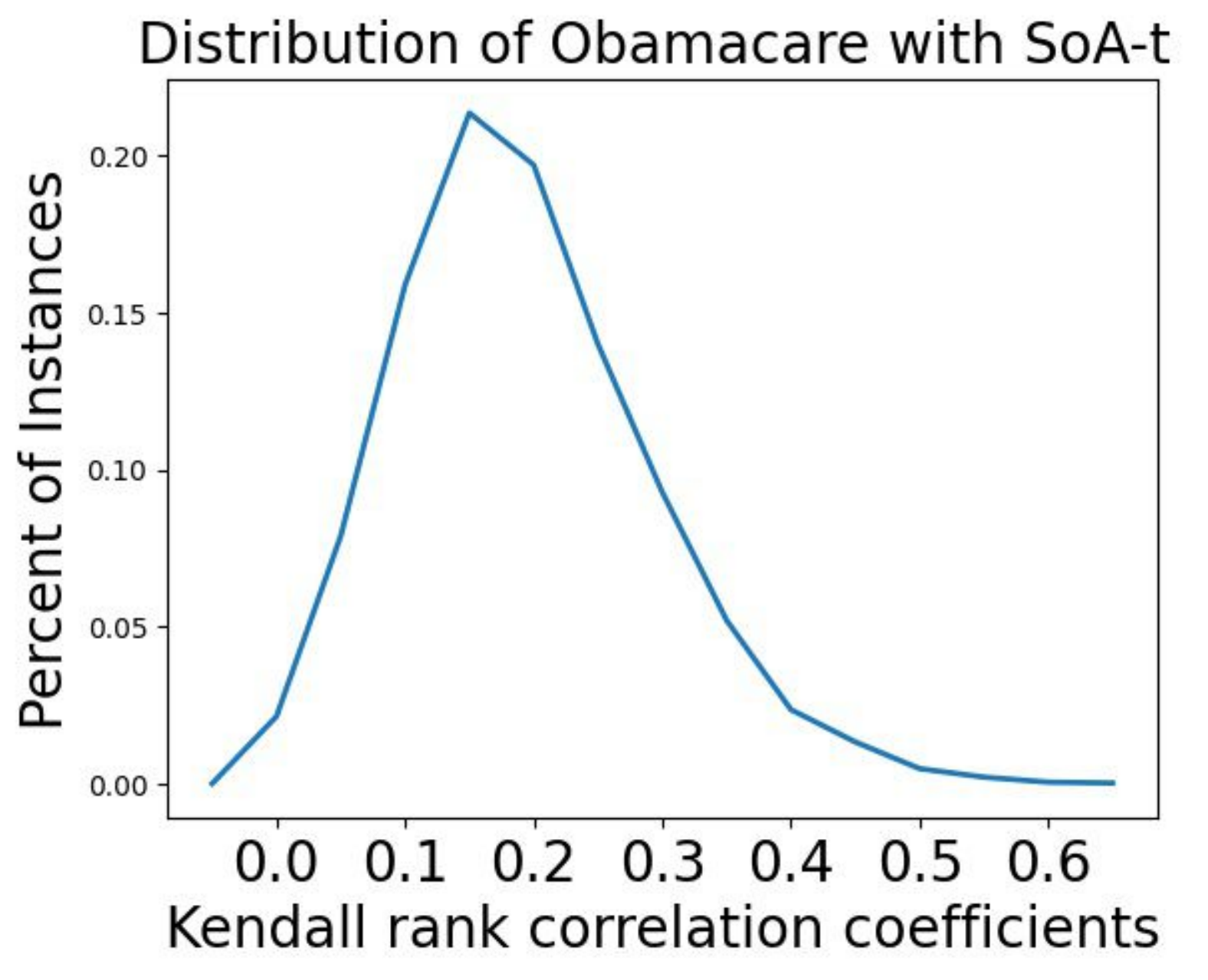}
        \caption{The distribution of Kendall's $\tau$ for different topics and ground truth datasets}
        \label{fig:distribution_all}
    \end{figure*}

\section{Hierarchical Clustering}
\setcounter{figure}{0}
    \label{sct:cluster}
\begin{figure*}[!hbt]
    \centering
    \includegraphics[width=1.00\columnwidth]{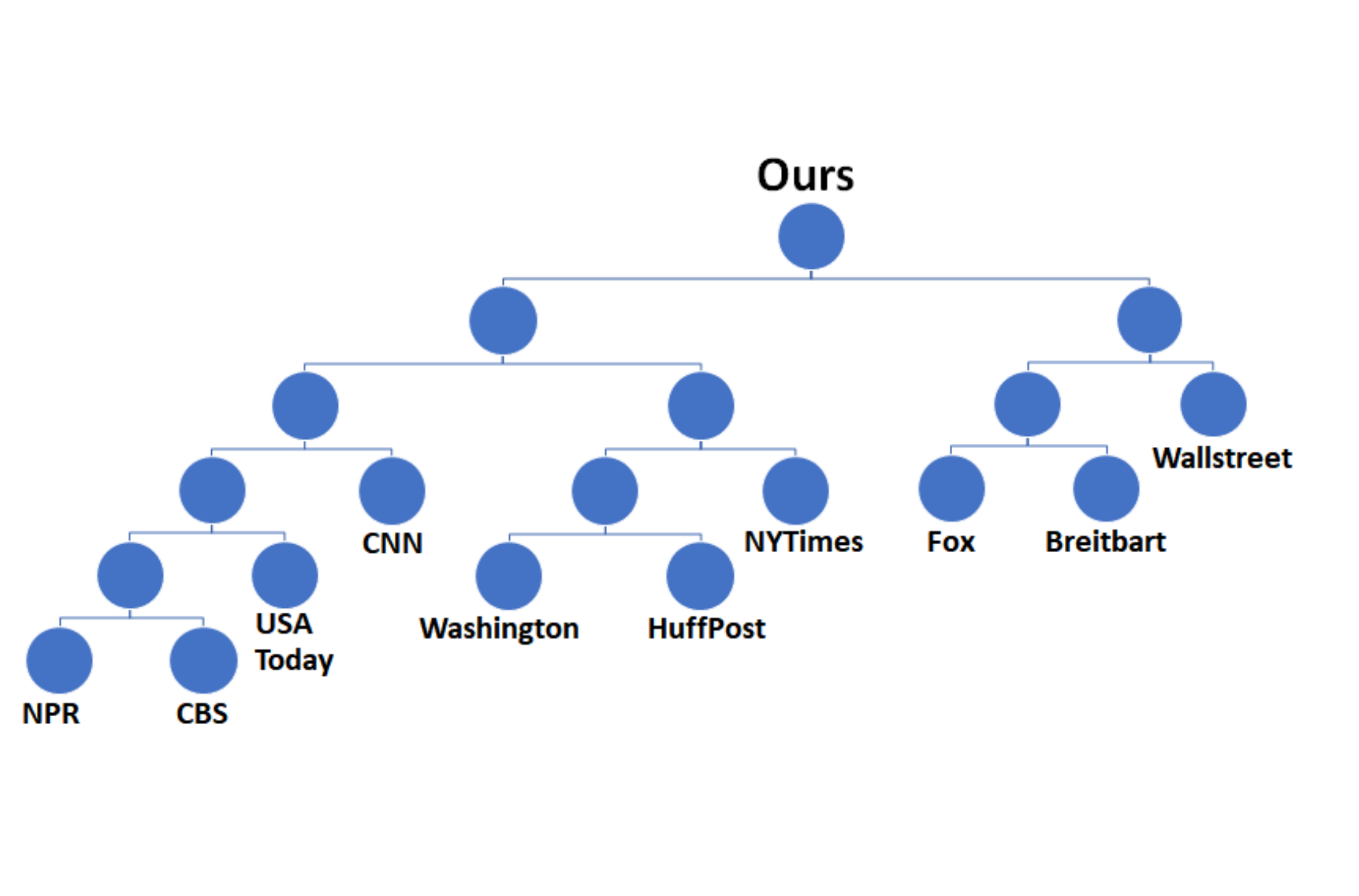}
    \includegraphics[width=1.00\columnwidth]{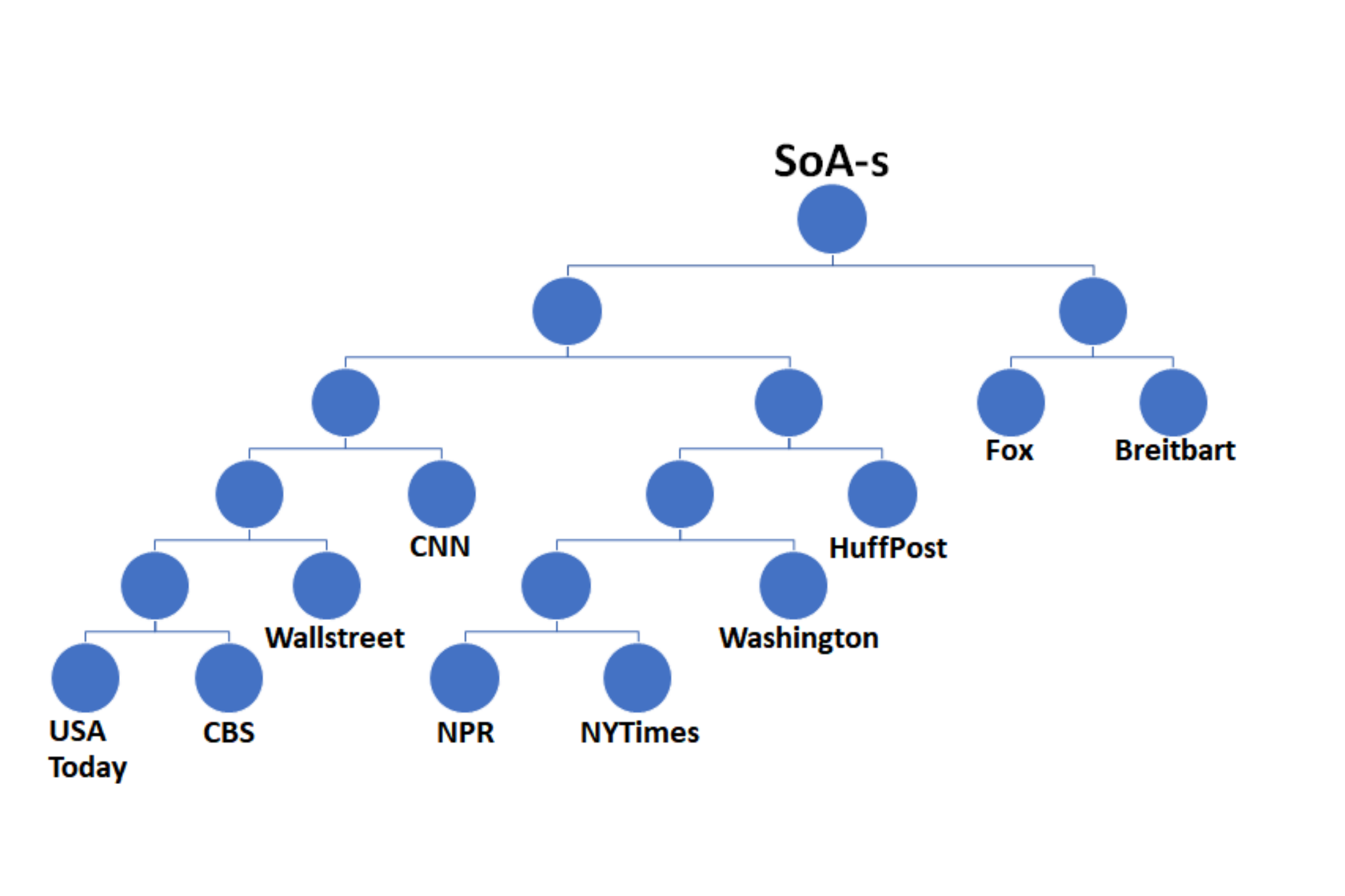}
    \caption{Hierarchical clustering result of our framework (left) and the SoA-s dataset (right). ``NYTimes'' : ``New York Times'', ``Wallstreet'' : ``The Wall Street Journal'', ``Washington'' : ``Washington Post''. The Kendall's $\tau$ is 0.41}
    \label{fig:hierarcical_clustering}
\end{figure*}
        In Figure \ref{fig:hierarcical_clustering}, we show the agglomerative hierarchical clustering result on media bias encodings generated using our framework and the SoA-source ground truth data for the topic ``The Affordable Care Act''. We can see that the grouping of outlets is similar for both of the clusterings, with the notable exception of  ``Wall Street Journal''. Though ``Wall Street Journal'' is placed at the center of the SoA-source dataset, our method places it closer to the political right, along with Fox and Breitbart. This is interesting as ``Wall Street Journal'' was recently bought by the parent company of Fox, the "News Corp", moving it to the ideological right. It is possible that the shift in the audience of ``Wall Street Journal'' (as captured by SoA-source) is lagging. Though these statements are purely speculative.

\end{document}